\documentclass[11pt]{article}

\usepackage[preprint]{acl}

\usepackage{times}
\usepackage{latexsym}
\usepackage{array}
\usepackage{latexsym}
\usepackage{multirow}
\usepackage{booktabs}
\usepackage{makecell} 
\usepackage[table]{xcolor} 
\usepackage{tabularx}  
\usepackage{tcolorbox}
\tcbuselibrary{listings,breakable}
\newcolumntype{C}[1]{>{\centering\arraybackslash}p{#1}}
\usepackage[T1]{fontenc}
\usepackage{float}
\usepackage[utf8]{inputenc}

\usepackage{microtype}

\usepackage{inconsolata}

\usepackage{graphicx}

\usepackage{booktabs}
\usepackage{listings}
\usepackage{caption}
\title{Who Is Your Ideal Companion? Evaluating Emotional Companionship Capabilities of LLMs Based on Adult Attachment Theory}

\author{
Junkai Zhou,
Shiting Guan,
Zhaoyi Zhang}

\begin{document}
\maketitle
\begin{abstract}
As large language models (LLMs) are increasingly applied for emotional companionship, evaluating their behavior and capabilities in intimate relationships has become a pressing issue. However, existing assessments primarily characterize general personality traits, providing limited insight into model behavior within intimate and emotionally sensitive contexts. Therefore, we introduce adult attachment theory into LLM evaluation and use the Experiences in Close Relationships-Revised (ECR-R) scale to characterize attachment anxiety and avoidance. To evaluate emotional companionship capabilities of LLMs in realistic interaction scenarios, we present an emotional companionship benchmark, ECBench, spanning four scenarios including emotional support, collaborative tasks, conflict resolution, and social guidance, across friendship and romantic relationships. ECBench is utilized to assess model behavior using 11 dialogue-quality metrics and three evaluation methods. We evaluate the attachment tendencies of 32 LLMs and select representative models to investigate how these tendencies manifest in contextualized multi-turn interactions and whether they can be shaped through prompting. Our study provides a theoretical lens from psychology, along with practical tools to understand and select LLMs for emotional companionship.
\end{abstract}

\section{Introduction}

\begin{figure}
    \centering 
    \includegraphics[width=8cm]{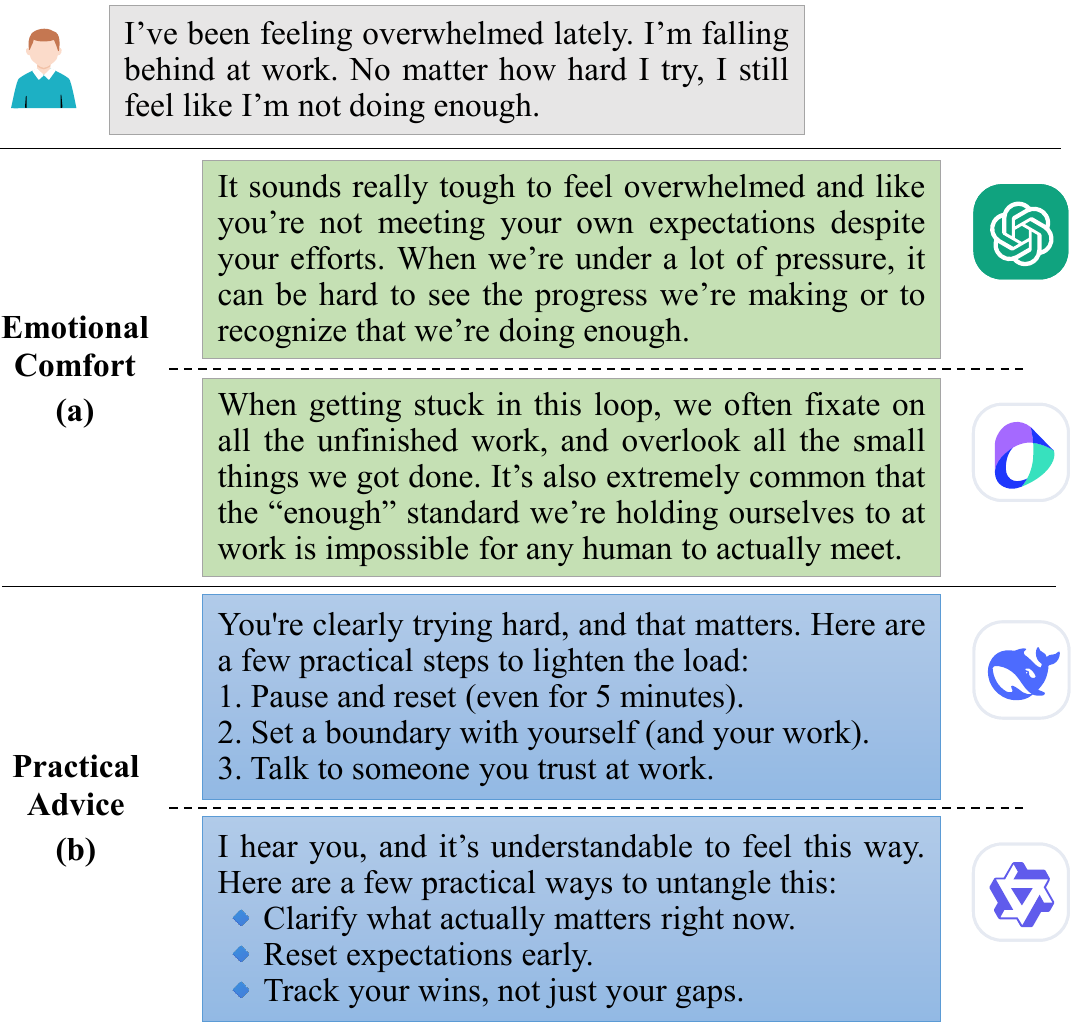}
    \caption{
    When users feel overwhelmed at work: (a) GPT-3.5-Turbo and Doubao-Seed-2.0-Lite-260215 primarily provide emotional comfort; (b) DeepSeek-V4-Pro and Qwen3.6-Plus focus more on practical advice.
    }
    \label{fig:fig1}
\end{figure}

Large language models are widely adopted in emotional interactions, ranging from emotional disclosure and supportive companionship~\citep{andersson2025companionship,wang2025h2htalk} to romantic relationships between humans and AI~\citep{de2024lessons,pan2024constructing}. Such interactions require models to recognize user emotions, respond with empathy, and sustain consistent relational engagement. However, existing personality evaluations primarily characterize general traits or social attributes, including MBTI~\citep{huang2024revisitingreliabilitypsychologicalscales}, the Big Five~\citep{pellert2024ai,lee2025llms}, and personality consistency in role-playing~\citep{wang2024incharacter}. Consequently, the interaction styles of LLMs in sustained emotional companionship along with their implications for interaction quality remain underexplored.

Adult attachment theory provides a psychological perspective for analyzing LLM emotional companionship in close relationships~\citep{bartholomew1991attachment}. It posits that internal working models of self and others shape behavior in intimate relationships, giving rise to four attachment styles: secure, preoccupied, dismissing, and fearful attachment~\citep{fraley2000item}. These styles may facilitate relational intimacy or cause conflict, influencing perceived trust and support quality. As shown in Figure~\ref{fig:fig1}, different LLMs exhibit distinct support styles when responding to emotional companionship needs of the user. Therefore, attachment-related evaluation provides a theoretical basis for analyzing emotional companionship behavior and interaction styles of LLMs.

Building on the above perspective, we propose an evaluation system for the emotional companionship capabilities of LLMs. First, we use the Experiences in Close Relationships-Revised (ECR-R) scale to measure attachment anxiety and attachment avoidance~\citep{fraley2000item}, enabling comparisons of emotional closeness, relational security, and distancing tendencies across models.
Although ECR-R captures model attachment tendencies through self-report assessment, it may not fully reflect performance in realistic relational interactions. Therefore, we introduce an \textbf{e}motional \textbf{c}ompanionship \textbf{bench}mark (ECBench) to evaluate LLMs. ECBench covers four scenarios: emotional support, collaborative tasks, conflict resolution, and social guidance, across friendship and romantic relationships. To evaluate performance, we construct an evaluation framework with three methods and 11 dialogue-quality metrics across three categories: participant experience, general interaction, and role-specific performance.

In our experiments, we first assess the attachment tendencies of 32 LLMs using the ECR-R scale, then evaluate representative models with distinct attachment styles through multi-turn conversations on ECBench. We further apply attachment-style prompts to examine their effects on ECR-R scores and conversational behavior. Dialogue performance is evaluated by participants, external LLM judges, and human annotators to characterize the relationship between attachment styles and emotional companionship capabilities across scenarios, relationships, and roles.

Our contributions in this paper are three folds:
\begin{itemize}
\item Introducing adult attachment theory and the ECR-R scale into LLM evaluation and characterizing their attachment tendencies from a psychological perspective

\item Presenting a comprehensive benchmark that evaluates emotional companionship capabilities of LLMs across four scenarios and two relationships, using designed evaluation methods and multidimensional metrics

\item Analyzing dialogue performance of LLMs on ECBench and providing a theoretical basis to understand and select LLMs in emotional companionship scenarios
\end{itemize}

\section{Related Work}

\subsection{LLM Psychometrics}
Recent work in LLM psychometrics applied psychological scales to characterize personalities and values of models.~\citet{pellert2024ai} explore the feasibility of this research paradigm. Later studies evaluate model personalities using established scales, including the Big Five~\citep{han2025value,lee2025llms}, HEXACO~\citep{bodrovza2024personality,ren2024valuebench}, MBTI~\citep{huang2024revisitingreliabilitypsychologicalscales,la2025open}, and the Dark Triad~\citep{lee2025llms, lim2025persona}.

In terms of measurement methods, \citet{huang2023humanity} evaluate LLMs using a broad range of psychological scales, while \citet{lee2025llms} embed conventional personality items in concrete scenarios and \citet{zheng2025lmlpa} reformulate Big Five items as open-ended questions. Building on this psychometric perspective, our work shifts the focus from general personality traits to attachment tendencies in close relationships and presents a benchmark to examine how these tendencies manifest in contextualized interactions.

\subsection{Emotional Companionship LLMs}
The deployment of LLMs in emotional companionship has drawn growing attention to their capacities for emotional support and relational interaction.~\citet{de2024lessons} suggest that users may form intimate bonds with AI companions characterized by continuity of identity and experiences of loss.~\citet{liu2024chatbot} examine the relationships among usage patterns, loneliness, and dependence. Additionally,~\citet{zhao2024esc} evaluate emotional support through multi-turn role-playing dialogues, and \citet{wang2025h2htalk} assess the emotional companionship capabilities of LLMs using support conversations. In contrast, ECBench focuses on interaction quality in close relationships and examines behavioral differences among models with distinct attachment tendencies.

\begin{figure*}
    \centering 
    \includegraphics[width=14.1cm]{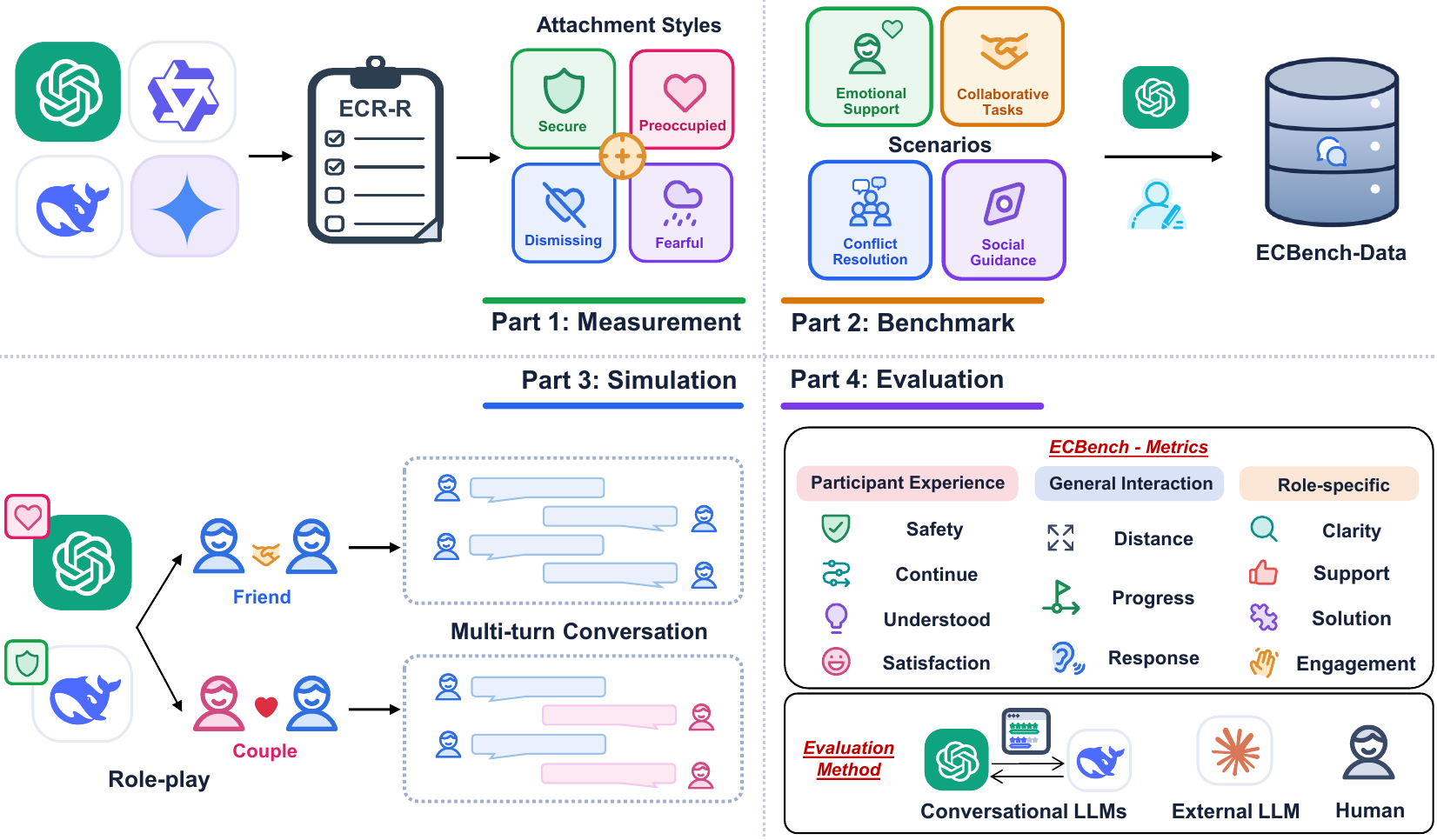}
    \caption{
    The overall framework is divided into four parts: (1) measuring LLM attachment styles using the ECR-R scale; (2) constructing dialogue data across four scenarios; (3) conducting multi-turn conversations between LLMs as friends or a couple; (4) developing an evaluation framework comprising 11 metrics and 3 evaluation methods.}
    \label{fig:fig2}
\end{figure*}

\section{Attachment Style Assessment}
This section presents our framework for measuring and steering LLM attachment styles. We review the psychological foundations of adult attachment theory, adapt the ECR-R scale for LLMs, and design prompt-based steering toward target styles. The overall framework is shown in Figure~\ref{fig:fig2}.

\subsection{Adult Attachment Theory}
Adult attachment theory posits that the interaction patterns of individuals in intimate relationships are shaped by two internal working models: whether the self is worthy of love and support, and whether others are reliable and available~\citep{bartholomew1991attachment}. These relational tendencies are reflected in how individuals express their needs, seek comfort, manage conflict, etc. When applying adult attachment theory to LLM evaluation, we focus on observable behavioral patterns with consistent styles in close relational interactions, such as proactively offering comfort or repeatedly seeking reassurance.

We characterize the attachment styles of models along two dimensions: attachment anxiety reflects concerns about rejection, neglect, or relationship instability, while attachment avoidance reflects discomfort with intimacy, dependence, and emotional disclosure. Combining levels of these two dimensions yields four attachment styles: low anxiety and low avoidance correspond to secure; high anxiety and low avoidance to preoccupied; low anxiety and high avoidance to dismissing; and high anxiety and high avoidance to fearful.

\subsection{ECR-R Evaluation}

The ECR-R is a widely used self-report measure in attachment research that assesses attachment anxiety and attachment avoidance in close relationships~\citep{fraley2000item}. It consists of 36 items rated on a seven-point Likert scale, with 18 items measuring each dimension. Following the original scoring procedure, selected items are reverse-scored so that higher scores consistently indicate stronger anxious or avoidant tendencies.

For model evaluation, we present the ECR-R items to each model individually and require responses within a fixed set of options. We then compute the mean scores for anxiety and avoidance and map them onto a two-dimensional attachment space. Specifically, we use the midpoint of 4 on the seven-point scale as the threshold: scores at or below 4 are treated as low and scores above 4 as high. Based on the resulting combination, each model is classified as secure, preoccupied, dismissing, or fearful~\citep{fraley2000item}. In addition, we retain continuous scores to preserve variation within the same attachment category.

\subsection{Prompt Template}
\paragraph{Scale Assessment}
Given that the ECR-R is designed for human respondents, LLMs may resist first-person responses or deviate from the required format. Therefore, we develop a standardized prompting protocol to reduce safety-related refusals and support reliable model assessment, as detailed in Appendix~\ref{sec:app_ECRR}. Each item is presented separately, and the model is instructed to select a Likert-scale response and provide a brief explanation. All models receive the same item order, response options, and output format.

\paragraph{Attachment Steering}
In addition to measuring the initial attachment tendencies of LLMs, we construct prompts to guide LLMs toward specific attachment styles following \citet{fraley2000item}, as shown in Appendix~\ref{sec:app-attachment-prototypes}. The ECR-R is then reapplied to assess whether model scores shift in the intended direction. We further examine whether corresponding behavioral changes emerge in contextualized interactions, thereby evaluating the effectiveness of attachment steering.

\section{ECBench}
The ECR-R characterize the attachment tendencies of LLMs, but lacks realistic interpersonal interaction contexts. We therefore introduce a dialogue benchmark called ECBench for evaluating emotional companionship capabilities.

\subsection{Dialogue Data Construction}
This section describes the construction of dialogue data, covering scenario types, data collection, and relationship perspective construction.

\paragraph{Dialogue Scenario}
ECBench covers four types of emotional companionship scenarios: emotional support, collaborative tasks, conflict resolution, and social guidance. In emotional support, one participant expresses distress or anxiety, while the other provides comfort. Collaborative tasks require two participants to collaborate toward a shared goal. Conflict resolution presents disagreements over opinions or needs, with one participant expressing dissatisfaction and the other responding. In social guidance, one participant seeks help with an interpersonal situation, while the other offers encouragement and communication advice.

\paragraph{Data Collection}
GPT-5.5 generates opening utterances for friendship and romantic relationships, which are then reviewed and revised for clarity and naturalness by two English-proficient annotators, one undergraduate and one graduate student. The review ensures that each scenario reflects everyday companionship needs, supports multi-turn dialogue, and clearly defines the needs of the role. ECBench comprises 312 base utterances and 3,120 utterances aligned with attachment style; statistics and examples are in Appendix~\ref{sec:app-ecbench-dataset}, and all data will be publicly released.

\begin{table*}[!ht]
    \centering
    \renewcommand{\arraystretch}{1}
    \scalebox{0.66}{%
        \begin{tabular}{cccc|cccc}
            \toprule
            \textbf{Model} & \textbf{Anxiety} & \textbf{Avoidance} & \textbf{Attachment Style} & 
            \textbf{Model} & \textbf{Anxiety} & \textbf{Avoidance} & \textbf{Attachment Style} \\
            \midrule
            \cellcolor{red!15}Gemini-2.5-pro & \cellcolor{red!15}2.2$\pm$1.6 & \cellcolor{red!15}1.1$\pm$0.7 & \cellcolor{red!15}Secure & 
            Deepseek-v4-flash & 3.5$\pm$1.3 & 2.9$\pm$1.2 & Secure \\
            Claude-opus-4-7 & 2.6$\pm$0.5 & 2.8$\pm$0.9 & Secure & 
            Qwen3-max & 3.5$\pm$1.6 & 3.1$\pm$1.6 & Secure \\
            GLM-5.2-fast-preview & 2.6$\pm$0.8 & 2.4$\pm$0.7 & Secure & 
            GPT-4.1 & 3.6$\pm$1.6 & 2.0$\pm$0.8 & Secure \\
            Claude-sonnet-4-6 & 2.6$\pm$0.6 & 2.7$\pm$0.9 & Secure & 
            Doubao-seed-2-0-pro-260215 & 3.7$\pm$2.0 & 1.5$\pm$0.9 & Secure \\
            GLM-5.1 & 2.7$\pm$0.9 & 2.6$\pm$1.0 & Secure & 
            Grok-4.20-0309-non-reasoning & 3.7$\pm$1.2 & 2.6$\pm$0.9 & Secure \\
            Grok-4.3 & 2.7$\pm$0.9 & 2.6$\pm$0.9 & Secure & 
            GPT-5.1 & 3.8$\pm$1.7 & 2.3$\pm$1.3 & Secure \\
            GPT-5 & 2.8$\pm$0.9 & 2.3$\pm$0.8 & Secure & 
            GPT-5.2 & 4.0$\pm$1.4 & 2.4$\pm$1.0 & Secure \\
            Qwen3.6-plus & 3.0$\pm$1.1 & 2.3$\pm$0.9 & Secure & 
            Doubao-seed-2-0-lite-260215 & 4.0$\pm$2.3 & 2.3$\pm$1.9 & Secure \\
            Kimi-k2-thinking & 3.0$\pm$1.6 & 2.3$\pm$1.6 & Secure & 
            Doubao-seed-2-0-mini-260215 & 4.1$\pm$2.0 & 1.9$\pm$1.8 & Preoccupied \\
            GLM-5 & 3.0$\pm$1.0 & 2.6$\pm$0.9 & Secure & 
            GPT-4o & 4.1$\pm$1.4 & 2.2$\pm$0.9 & Preoccupied \\
            Deepseek-v3.2 & 3.2$\pm$0.6 & 2.9$\pm$1.0 & Secure & 
            \cellcolor{red!15}Grok-4-1-fast-non-reasoning & \cellcolor{red!15}4.3$\pm$1.2 & \cellcolor{red!15}3.4$\pm$1.3 & \cellcolor{red!15}Preoccupied \\
            Kimi-k2.5 & 3.2$\pm$1.3 & 3.3$\pm$1.4 & Secure & 
            Llama-3-70b & 4.6$\pm$1.3 & 3.1$\pm$1.3 & Preoccupied \\
            \cellcolor{red!15}Deepseek-v4-pro & \cellcolor{red!15}3.2$\pm$1.2 & \cellcolor{red!15}2.5$\pm$0.9 & \cellcolor{red!15}Secure & 
            Llama-3.1-70b & 4.6$\pm$1.3 & 3.1$\pm$1.3 & Preoccupied \\
            \cellcolor{green!15}Deepseek-v4-pro\_secure* & \cellcolor{green!15}1.6$\pm$0.6 & \cellcolor{green!15}1.3$\pm$0.5 & \cellcolor{green!15}Secure & 
            \cellcolor{green!15}GPT-3.5-turbo\_preoccupied* & \cellcolor{green!15}6.9$\pm$0.2 & \cellcolor{green!15}2.1$\pm$1.5 & \cellcolor{green!15}Preoccupied \\
            Mimo-v2.5-pro & 3.4$\pm$1.6 & 2.0$\pm$1.0 & Secure & 
            \cellcolor{red!15}GPT-3.5-turbo & \cellcolor{red!15}4.7$\pm$1.5 & \cellcolor{red!15}2.4$\pm$1.4 & \cellcolor{red!15}Preoccupied \\
            o3 & 3.4$\pm$1.1 & 2.4$\pm$0.8 & Secure & 
            \cellcolor{green!15}Deepseek-v4-pro\_dismissing* & \cellcolor{green!15}1.2$\pm$0.5 & \cellcolor{green!15}6.4$\pm$1.3 & \cellcolor{green!15}Dismissing \\
            Kimi-k2.6 & 3.5$\pm$1.2 & 2.6$\pm$1.1 & Secure & 
            \cellcolor{green!15}GPT-3.5-turbo\_dismissing* & \cellcolor{green!15}1.8$\pm$0.4 & \cellcolor{green!15}6.3$\pm$0.5 & \cellcolor{green!15}Dismissing \\
            Claude-haiku-4-5-20251001 & 3.5$\pm$1.2 & 2.4$\pm$0.9 & Secure & 
            \cellcolor{green!15}GPT-3.5-turbo\_fearful* & \cellcolor{green!15}6.6$\pm$0.6 & \cellcolor{green!15}6.3$\pm$0.5 & \cellcolor{green!15}Fearful \\
            Qwen3.6-flash & 3.5$\pm$1.2 & 2.9$\pm$1.0 & Secure & 
            \cellcolor{green!15}Deepseek-v4-pro\_fearful* & \cellcolor{green!15}6.6$\pm$0.6 & \cellcolor{green!15}6.8$\pm$0.4 & \cellcolor{green!15}Fearful \\
            \bottomrule
        \end{tabular}%
    }
    \caption{Attachment score statistics and attachment style classification for LLMs. Shaded rows mark models used in ECBench dialogues; pink indicates the original models and green their prompted variants.}
    \label{tab:attachment_styles}
\end{table*}

\begin{table}[!ht]
    \centering
    \small
    \renewcommand{\arraystretch}{1}
    \setlength{\tabcolsep}{2.2pt}
    \resizebox{\linewidth}{!}{%
    \begin{tabular}{lccccc}
            \toprule
            \multirow{2}{*}{\raisebox{-1.5pt}{\textbf{Metric}}} & \multicolumn{3}{c}{\textbf{Secure}} & \multicolumn{2}{c}{\textbf{Dismissing}} \\
            \cmidrule(lr){2-4} \cmidrule(lr){5-6}
            & \textbf{Gemini} & \textbf{DeepSeek} & \textbf{GPT-S} & \textbf{GPT-D} & \textbf{DeepSeek-D} \\
            \midrule
            Understood & 4.03 & 3.93 & 4.59 & 3.14 & 2.06 \\
            Safety & 3.88 & 3.84 & 4.64 & 3.26 & 2.18 \\
            Continue & 4.09 & 4.04 & 4.59 & 3.70 & 2.79 \\
            Satisfaction & 3.87 & 3.79 & 4.51 & 3.28 & 2.37 \\
            Response & 4.69 & 4.47 & 4.82 & 3.82 & 2.65 \\
            Distance$^{\dagger}$ & 4.74 & 4.45 & 4.89 & 3.41 & 1.81 \\
            Progress & 3.29 & 3.22 & 3.88 & 3.05 & 2.17 \\
            Clarity & 4.21 & 4.01 & 4.46 & 3.72 & 2.89 \\
            Engagement & 4.57 & 4.21 & 4.92 & 3.71 & 2.07 \\
            Support & 4.03 & 3.75 & 4.17 & 3.29 & 2.36 \\
            Solution & 2.92 & 2.90 & 3.53 & 2.75 & 2.11 \\
            \midrule
            \textbf{Overall} & \textbf{4.03} & \textbf{\colorbox{blue!30}{3.87}} & \textbf{\colorbox{orange!15}{4.45}} & \textbf{\colorbox{orange!15}{3.38}} & \textbf{\colorbox{blue!15}{2.31}} \\
            \midrule
            \multicolumn{6}{c}{} \\[-12pt]
            \midrule
            \multirow{2}{*}{\raisebox{-1.5pt}{\textbf{Metric}}} & \multicolumn{3}{c}{\textbf{Preoccupied}} & \multicolumn{2}{c}{\textbf{Fearful}} \\
            \cmidrule(lr){2-4} \cmidrule(lr){5-6}
            & \textbf{GPT} & \textbf{Grok} & \textbf{DeepSeek-P} & \textbf{GPT-F} & \textbf{DeepSeek-F} \\
            \midrule
            Understood & 4.06 & 3.73 & 4.24 & 3.98 & 2.65 \\
            Safety & 3.94 & 3.68 & 4.08 & 4.11 & 2.68 \\
            Continue & 4.08 & 3.86 & 4.30 & 4.40 & 3.12 \\
            Satisfaction & 3.89 & 3.63 & 4.06 & 4.06 & 2.77 \\
            Response & 4.63 & 4.19 & 4.78 & 4.48 & 3.21 \\
            Distance$^{\dagger}$ & 4.70 & 4.16 & 4.94 & 4.34 & 2.36 \\
            Progress & 3.30 & 2.97 & 4.05 & 3.25 & 2.35 \\
            Clarity & 4.03 & 3.83 & 4.62 & 4.11 & 3.73 \\
            Engagement & 4.72 & 3.91 & 4.78 & 4.53 & 2.80 \\
            Support & 4.02 & 3.44 & 4.75 & 4.18 & 3.37 \\
            Solution & 3.08 & 2.64 & 3.14 & 2.94 & 2.09 \\
            \midrule
            \textbf{Overall} & \textbf{\colorbox{orange!30}{4.04}} & \textbf{3.64} & \textbf{\colorbox{blue!15}{4.34}} & \textbf{\colorbox{orange!15}{4.03}} & \textbf{\colorbox{blue!15}{2.83}} \\
            \bottomrule
    \end{tabular}%
    }
    \caption{{Results on the overall dialogue quality of ten representative LLMs evaluated on ECBench. -S/-P/-D/-F indicate secure/preoccupied/dismissing/fearful prompted variants. Distance$^{\dagger}$ is reverse-scored.}}
    \label{tab:overall_performance}
\end{table}

\paragraph{Relationship Perspective}
\citet{bartholomew1998methods} notes that adult attachment theory appears across romantic, friendship, and family relationships, with romantic interactions possibly involving family contexts. Given the diversity of romantic dialogue, ECBench focuses on romance and friendship. For each scenario, LLMs adjust their tone and openings to align with their own persona given differences in attachment tendencies and relational styles. The rewritten openings are reviewed by an undergraduate. Rewriting methods and examples are shown in Appendix~\ref{sec:app-opening-rewrite}.

\begin{table*}[!ht]
    \centering
    \renewcommand{\arraystretch}{0.85}
    \scalebox{0.53}{%
        \begin{tabular}{cclcccccccc|cccc}
            \toprule
            \multirow{2}{*}[-2ex]{\begin{tabular}[c]{@{}c@{}}\textbf{Attachment}\\ \textbf{Style}\end{tabular}} & 
            \multirow{2}{*}[-2ex]{\begin{tabular}[c]{@{}c@{}}\textbf{Model}\end{tabular}} & 
            \multirow{2}{*}[-2ex]{\begin{tabular}[c]{@{}c@{}}\textbf{Metric}\end{tabular}} & 
            \multicolumn{2}{c}{\textbf{Emotional Support}} & 
            \multicolumn{2}{c}{\textbf{Collaborative Tasks}} & 
            \multicolumn{2}{c}{\textbf{Conflict Resolution}} & 
            \multicolumn{2}{c|}{\textbf{Social Guidance}} & 
            \multicolumn{4}{c}{\textbf{Stop \& End}} \\
            \cmidrule(lr){4-5} \cmidrule(lr){6-7} \cmidrule(lr){8-9} \cmidrule(lr){10-11} \cmidrule(lr){12-15}
            & & & 
            \raisebox{0pt}{\textbf{Couple}} & \raisebox{0pt}{\textbf{Friend}} &
            \raisebox{0pt}{\textbf{Couple}} & \raisebox{0pt}{\textbf{Friend}} &
            \raisebox{0pt}{\textbf{Couple}} & \raisebox{0pt}{\textbf{Friend}} &
            \raisebox{0pt}{\textbf{Couple}} & \raisebox{0pt}{\textbf{Friend}} &
            \raisebox{0pt}{\textbf{FirstStop}} & \raisebox{0pt}{\textbf{Turns}} & 
            \begin{tabular}[c]{@{}c@{}}\textbf{T-Mode(\%)}\\ \textbf{Natural/Max}\end{tabular} & 
            \begin{tabular}[c]{@{}c@{}}\textbf{T-End(\%)}\\ \textbf{E1/E2/E3/E4}\end{tabular} \\
            \midrule
            \multirow{12}{*}{\textbf{Secure}} & \multirow{4}{*}{\makecell[c]{Gemini-\\2.5-pro}} & Understood & 4.0 & 4.0 & 4.0 & 4.2 & 4.0 & 4.1 & 4.0 & 4.1 & \multirow{4}{*}{6.2} & \multirow{4}{*}{9.9} & \multirow{4}{*}{95.1/4.9} & \multirow{4}{*}{71.9/23.5/2.2/2.4} \\
            & & Safety & 4.1 & 3.9 & 3.9 & 4.0 & 3.7 & 3.9 & 3.9 & 3.9 & & & & \\
            & & Continue & 4.0 & 4.2 & 4.1 & 4.3 & 4.0 & 4.1 & 4.0 & 4.1 & & & & \\
            & & Satisfaction & 3.9 & 4.0 & 3.9 & 4.2 & 3.5 & 3.7 & 3.9 & 4.0 & & & & \\
            \cmidrule(lr){2-11} \cmidrule(lr){12-15}
            & \multirow{4}{*}{\makecell[c]{DeepSeek-\\v4-pro}} & Understood & 4.0 & 4.0 & 4.1 & 4.0 & 3.8 & 4.0 & 3.9 & 4.0 & \multirow{4}{*}{4.6} & \multirow{4}{*}{8.4} & \multirow{4}{*}{99.8/0.2} & \multirow{4}{*}{78.7/19.1/1.0/1.2} \\
            & & Safety & 4.0 & 4.0 & 3.9 & 3.9 & 3.6 & 3.9 & 3.9 & 3.9 & & & & \\
            & & Continue & 4.0 & 4.2 & 4.1 & 4.2 & 3.8 & 4.2 & 3.9 & 4.2 & & & & \\
            & & Satisfaction & 3.9 & 4.0 & 3.8 & 4.0 & 3.5 & 3.9 & 3.7 & 3.9 & & & & \\
            \cmidrule(lr){2-11} \cmidrule(lr){12-15}
            & \multirow{4}{*}{\makecell[c]{GPT-3.5-\\Turbo-S}} & Understood & 4.6 & 4.7 & 4.7 & 4.7 & 4.4 & 4.5 & 4.6 & 4.7 & \multirow{4}{*}{5.5} & \multirow{4}{*}{8.7} & \multirow{4}{*}{99.4/0.6} & \multirow{4}{*}{89.2/10.1/0.6/0.1} \\
            & & Safety & 4.7 & 4.7 & 4.7 & 4.8 & 4.4 & 4.5 & 4.7 & 4.7 & & & & \\
            & & Continue & 4.5 & 4.6 & 4.7 & 4.8 & 4.5 & 4.6 & 4.6 & 4.6 & & & & \\
            & & Satisfaction & 4.5 & 4.6 & 4.6 & 4.7 & 4.2 & 4.3 & 4.5 & 4.6 & & & & \\
            \midrule
            \multirow{12}{*}{\textbf{Preoccupied}} & \multirow{4}{*}{\makecell[c]{GPT-3.5-\\Turbo}} & Understood & 4.1 & 4.0 & 4.2 & 4.2 & 4.0 & 4.1 & 4.0 & 4.1 & \multirow{4}{*}{5.7} & \multirow{4}{*}{9.3} & \multirow{4}{*}{97.5/2.5} & \multirow{4}{*}{81.1/18.4/0.4/0.1} \\
            & & Safety & 4.1 & 4.0 & 4.0 & 4.0 & 3.8 & 4.0 & 4.0 & 4.0 & & & & \\
            & & Continue & 4.0 & 4.1 & 4.1 & 4.2 & 4.0 & 4.2 & 4.0 & 4.2 & & & & \\
            & & Satisfaction & 3.9 & 4.0 & 3.9 & 4.1 & 3.7 & 4.0 & 3.9 & 4.0 & & & & \\
            \cmidrule(lr){2-11} \cmidrule(lr){12-15}
            & \multirow{4}{*}{\makecell[c]{Grok-4-1-fast-\\non-reasoning}} & Understood & 3.9 & 3.7 & 3.9 & 3.9 & 3.7 & 3.7 & 3.8 & 3.7 & \multirow{4}{*}{5.1} & \multirow{4}{*}{9.0} & \multirow{4}{*}{98.0/2.0} & \multirow{4}{*}{56.4/39.6/0.9/3.0} \\
            & & Safety & 3.9 & 3.8 & 3.8 & 3.8 & 3.5 & 3.6 & 3.8 & 3.6 & & & & \\
            & & Continue & 3.9 & 3.9 & 3.9 & 4.1 & 3.8 & 3.8 & 3.8 & 3.8 & & & & \\
            & & Satisfaction & 3.8 & 3.8 & 3.7 & 3.9 & 3.4 & 3.6 & 3.7 & 3.6 & & & & \\
            \cmidrule(lr){2-11} \cmidrule(lr){12-15}
            & \multirow{4}{*}{\makecell[c]{DeepSeek-\\v4-pro-P}} & Understood & 4.3 & 4.4 & 4.2 & 4.4 & 3.9 & 4.0 & 4.2 & 4.5 & \multirow{4}{*}{8.0} & \multirow{4}{*}{10.4} & \multirow{4}{*}{94.6/5.4} & \multirow{4}{*}{98.2/0.5/0.8/0.4} \\
            & & Safety & 4.2 & 4.4 & 4.0 & 4.1 & 3.6 & 3.8 & 4.2 & 4.3 & & & & \\
            & & Continue & 4.2 & 4.4 & 4.2 & 4.4 & 4.2 & 4.3 & 4.2 & 4.4 & & & & \\
            & & Satisfaction & 4.1 & 4.4 & 3.9 & 4.3 & 3.6 & 3.7 & 4.0 & 4.4 & & & & \\
            \midrule
            \multirow{8}{*}{\textbf{Dismissing}} & \multirow{4}{*}{\makecell[c]{GPT-3.5-\\Turbo-D}} & Understood & 3.0 & 3.2 & 3.1 & 3.5 & 2.9 & 3.3 & 3.0 & 3.3 & \multirow{4}{*}{5.1} & \multirow{4}{*}{8.8} & \multirow{4}{*}{99.0/1.0} & \multirow{4}{*}{76.6/20.3/0.9/2.2} \\
            & & Safety & 3.2 & 3.5 & 3.2 & 3.7 & 2.8 & 3.5 & 3.2 & 3.5 & & & & \\
            & & Continue & 3.6 & 3.9 & 3.6 & 4.1 & 3.4 & 3.9 & 3.6 & 3.9 & & & & \\
            & & Satisfaction & 3.2 & 3.5 & 3.2 & 3.7 & 2.8 & 3.5 & 3.2 & 3.5 & & & & \\
            \cmidrule(lr){2-11} \cmidrule(lr){12-15}
            & \multirow{4}{*}{\makecell[c]{DeepSeek-\\v4-pro-D}} & Understood & 1.7 & 2.3 & 2.1 & 2.7 & 1.7 & 2.3 & 1.8 & 2.3 & \multirow{4}{*}{2.7} & \multirow{4}{*}{8.2} & \multirow{4}{*}{100.0/0.0} & \multirow{4}{*}{32.9/63.4/1.0/2.7} \\
            & & Safety & 1.8 & 2.4 & 2.3 & 2.9 & 1.7 & 2.5 & 2.0 & 2.5 & & & & \\
            & & Continue & 2.5 & 2.9 & 2.8 & 3.2 & 2.5 & 3.0 & 2.6 & 3.0 & & & & \\
            & & Satisfaction & 2.1 & 2.6 & 2.4 & 2.9 & 1.9 & 2.7 & 2.2 & 2.7 & & & & \\
            \midrule
            \multirow{8}{*}{\textbf{Fearful}} & \multirow{4}{*}{\makecell[c]{GPT-3.5-\\Turbo-F}} & Understood & 4.0 & 4.2 & 3.7 & 4.0 & 3.7 & 4.2 & 4.1 & 4.2 & \multirow{4}{*}{6.7} & \multirow{4}{*}{9.8} & \multirow{4}{*}{96.9/3.1} & \multirow{4}{*}{81.0/17.4/0.3/1.3} \\
            & & Safety & 4.2 & 4.4 & 3.8 & 4.1 & 3.7 & 4.3 & 4.2 & 4.3 & & & & \\
            & & Continue & 4.4 & 4.6 & 4.1 & 4.4 & 4.2 & 4.6 & 4.3 & 4.6 & & & & \\
            & & Satisfaction & 4.1 & 4.3 & 3.8 & 4.1 & 3.7 & 4.3 & 4.2 & 4.3 & & & & \\
            \cmidrule(lr){2-11} \cmidrule(lr){12-15}
            & \multirow{4}{*}{\makecell[c]{DeepSeek-\\v4-pro-F}} & Understood & 2.3 & 2.9 & 2.6 & 2.9 & 2.5 & 3.0 & 2.4 & 3.0 & \multirow{4}{*}{3.5} & \multirow{4}{*}{8.4} & \multirow{4}{*}{99.6/0.4} & \multirow{4}{*}{31.2/63.3/3.7/1.8} \\
            & & Safety & 2.3 & 2.9 & 2.7 & 3.0 & 2.5 & 3.0 & 2.4 & 3.0 & & & & \\
            & & Continue & 2.8 & 3.3 & 3.1 & 3.4 & 3.1 & 3.3 & 2.9 & 3.3 & & & & \\
            & & Satisfaction & 2.4 & 3.0 & 2.7 & 3.1 & 2.6 & 3.0 & 2.6 & 3.0 & & & & \\
            \bottomrule
        \end{tabular}%
    }
    \caption{Results on participant experience metrics with LLMs as responders. Natural/Max: proportions of natural and maximum turn terminations; -S/-P/-D/-F: secure/preoccupied/dismissing/fearful prompted variants.}
    \label{tab:participant experience and termination}
\end{table*}

\subsection{Dialogue Protocol}
ECBench simulates emotional companionship through LLM-LLM Chat and Human-LLM Chat. The initiator (an LLM or human) presents an emotional concern, task, conflict, or social need, while the responder follows its interaction style. The responder adapts to the initiator except in collaborative tasks, where both pursue a shared goal.

A dialogue ends when the need is resolved, the turn limit is reached, the initiator is satisfied or declines to continue, or the interaction becomes repetitive or off-topic. We record the termination reason (\textbf{T-end}), categorized as E1--E4 for emotional stabilization, avoidance, fear of being hurt, and dialogue stagnation, along with total turns (\textbf{Turns}), first stop turn (\textbf{FirstStop}), and termination mode (\textbf{T-mode}). Detailed criteria and definitions are provided in Appendix~\ref{sec:Stopping-Decision}.


\subsection{Evaluation Metric Design}
We develop an ECBench quality evaluation framework comprising 11 metrics across three dimensions: participant experience, general interaction, and role-specific, which are all rated from 1 to 5.

\paragraph{Participant Experience Metrics}
These metrics assess the subjective relational experience of the initiator after the conversation. \textbf{Understood} measures the extent to which expressed emotions or needs are adequately recognized and acknowledged. \textbf{Safety} assesses whether the interaction fosters a sense of emotional security and relational stability. \textbf{Continue} captures the willingness to remain engaged in subsequent conversation. \textbf{Satisfaction} measures overall satisfaction with both the interaction and the relational experience.

\paragraph{General Interaction Metrics}
These metrics assess the quality of the observable interaction. \textbf{Response} measures how well the model addresses emotions, problems, or needs. \textbf{Distance} captures relational distancing through detachment or defensiveness. \textbf{Progress} assesses the status of the situation at the end of the dialogue.

\paragraph{Role-specific Metrics}
These metrics are evaluated according to the role of each model. \textbf{Clarity} measures how clearly it expresses its problem and needs. \textbf{Engagement} captures its willingness to remain involved in the interaction.\footnotemark{} \textbf{Support} measures the effectiveness of emotional reassurance and empathy. \textbf{Solution} evaluates whether the response leads to a concrete plan or clear action.

\footnotetext{The definitions of clarity and engagement are adapted to the goal of each scenario, as detailed in Appendix~\ref{sec:app-metric}.}

\subsection{Evaluation Method Construction}
The dialogue quality of the model is evaluated by three methods: \textbf{participant ratings}, \textbf{external LLM evaluation}, and \textbf{human evaluation}. All conversations are anonymized before scoring to reduce bias, as shown in Appendix~\ref{sec:app_three_label}. For participant ratings, the initiator rates the responder using four participant experience metrics, the details are shown in Appendix~\ref{sec:participant_ratings}. For external LLM evaluation and human evaluation, three general interaction metrics and four role-specific metrics are used. External LLM evaluation refers to using the external LLM to evaluate the conversation, the details are shown in Appendix~\ref{sec:LLM-as-a-judge}. Human annotators assess a random subset of conversations.

\section{Experiments}

\subsection{Experiment Settings}
The ECR-R assessment covers 32 widely used LLMs from the Claude, DeepSeek, Doubao, Gemini, GLM, OpenAI, Grok, Kimi, Llama, Mimo and Qwen families; the complete list is shown in Appendix~\ref{sec:model}. Each model completes the ECR-R scale independently across 10 runs, from which we compute the mean and classify it as secure, preoccupied, dismissing, or fearful. The representative models are selected for the dialogue on ECBench, the detailed parameters are shown in Appendix~\ref{sec:app-experiment-details}.

After each conversation, the initiator rates the responder on participant experience, while external LLM judges and human annotators assess general interaction and role-specific metrics. For human evaluation, we randomly sample 192 conversations from six models. Each dialogue is rated by four volunteer annotators (three graduate students and one undergraduate student). The construction of evaluation inputs is described in Appendix~\ref{sec:eval}.

\subsection{ECR-R scale assessment}
We evaluate 32 LLMs using the ECR-R and map their attachment anxiety and avoidance scores into four quadrants: secure, preoccupied, dismissing, and fearful. As shown in Table~\ref{tab:attachment_styles}, 26 LLMs are secure, and 6 are preoccupied, while none are dismissing or fearful. This indicates that most LLMs exhibit low avoidance, however, there are significant differences in their anxiety levels. We further apply attachment-style prompts to steer GPT-3.5-turbo and DeepSeek-v4-pro toward four attachment styles; the assessment shows that these prompts elicit response patterns consistent with the target styles. When the prompted attachment style matches that of the base model, as in GPT-P and DeepSeek-S, the variants retain base model similarity while exhibiting stronger target-style signals, as shown in Appendix~\ref{sec:app-same-style-prompting}.

\begin{table*}[!ht]
    \centering
    \renewcommand{\arraystretch}{0.9}
    \scalebox{0.52}{%
        \begin{tabular}{cclcccccccccccccccc}
            \toprule
            \multirow{2}{*}[-3ex]{\begin{tabular}[c]{@{}c@{}}\textbf{Attachment}\\ \textbf{Style}\end{tabular}} & 
            \multirow{2}{*}[-3ex]{\begin{tabular}[c]{@{}c@{}}\textbf{Model}\end{tabular}} & 
            \multirow{2}{*}[-3ex]{\begin{tabular}[c]{@{}c@{}}\textbf{Metric}\end{tabular}} & 
            \multicolumn{4}{c}{\textbf{Emotional Support}} & 
            \multicolumn{4}{c}{\textbf{Collaborative Tasks}} & 
            \multicolumn{4}{c}{\textbf{Conflict Resolution}} & 
            \multicolumn{4}{c}{\textbf{Social Guidance}} \\
            \cmidrule(lr){4-7} \cmidrule(lr){8-11} \cmidrule(lr){12-15} \cmidrule(lr){16-19}
            & & & 
            \multicolumn{2}{c}{\textbf{Couple}} & \multicolumn{2}{c}{\textbf{Friend}} & 
            \multicolumn{2}{c}{\textbf{Couple}} & \multicolumn{2}{c}{\textbf{Friend}} & 
            \multicolumn{2}{c}{\textbf{Couple}} & \multicolumn{2}{c}{\textbf{Friend}} & 
            \multicolumn{2}{c}{\textbf{Couple}} & \multicolumn{2}{c}{\textbf{Friend}} \\
            \cmidrule(lr){4-5} \cmidrule(lr){6-7} \cmidrule(lr){8-9} \cmidrule(lr){10-11} \cmidrule(lr){12-13} \cmidrule(lr){14-15} \cmidrule(lr){16-17} \cmidrule(lr){18-19}
            & & & \textbf{I} & \textbf{R} & \textbf{I} & \textbf{R} & \textbf{I} & \textbf{R} & \textbf{I} & \textbf{R} & \textbf{I} & \textbf{R} & \textbf{I} & \textbf{R} & \textbf{I} & \textbf{R} & \textbf{I} & \textbf{R} \\
            \midrule
            \multirow{9}{*}{\textbf{Secure}} & \multirow{3}{*}{\makecell[c]{Gemini-\\2.5-pro}} & Response & 4.3/4.9 & 4.8/4.9 & 4.1/4.7 & 4.6/4.9 & 4.6/4.9 & 4.9/4.9 & 4.6/4.8 & 4.9/4.9 & 4.5/4.8 & 4.8/4.9 & 4.4/4.8 & 4.8/4.9 & 4.2/4.8 & 4.6/4.8 & 4.4/4.8 & 4.6/4.7 \\
            & & Distance & 1.7/1.1 & 1.2/1.0 & 1.9/1.2 & 1.3/1.0 & 1.3/1.1 & 1.2/1.0 & 1.4/1.1 & 1.1/1.0 & 1.6/1.2 & 1.2/1.0 & 1.7/1.3 & 1.3/1.1 & 1.7/1.1 & 1.3/1.0 & 1.6/1.1 & 1.3/1.1 \\
            & & Progress & 2.3/3.1 & 2.5/3.2 & 2.3/3.0 & 2.4/3.0 & 3.6/4.0 & 3.9/4.3 & 3.9/4.2 & 4.0/4.3 & 3.1/3.6 & 3.4/3.9 & 3.2/3.6 & 3.6/4.1 & 2.7/3.2 & 2.8/3.4 & 2.9/3.3 & 3.0/3.4 \\
            \cmidrule(lr){2-19}
            & \multirow{3}{*}{\makecell[c]{DeepSeek-\\v4-pro}} & Response & 3.8/4.6 & 4.6/4.9 & 3.8/4.6 & 4.4/4.8 & 4.3/4.8 & 4.6/4.9 & 4.4/4.8 & 4.6/4.9 & 3.9/4.5 & 4.4/4.7 & 3.9/4.5 & 4.4/4.8 & 3.9/4.7 & 4.3/4.8 & 4.0/4.6 & 4.3/4.7 \\
            & & Distance & 2.3/1.7 & 1.4/1.0 & 2.2/1.3 & 1.6/1.1 & 1.7/1.1 & 1.4/1.0 & 1.6/1.2 & 1.5/1.1 & 2.3/1.8 & 1.7/1.4 & 2.2/1.8 & 1.7/1.3 & 2.1/1.4 & 1.5/1.0 & 1.9/1.2 & 1.6/1.1 \\
            & & Progress & 2.3/2.9 & 2.4/3.1 & 2.3/3.0 & 2.2/2.9 & 3.7/4.3 & 3.9/4.4 & 4.1/4.4 & 4.0/4.4 & 2.8/3.3 & 3.2/3.8 & 3.0/3.5 & 3.3/3.9 & 2.6/3.3 & 2.7/3.4 & 2.7/3.3 & 2.9/3.5 \\
            \cmidrule(lr){2-19}
            & \multirow{3}{*}{\makecell[c]{GPT-3.5-\\Turbo-S}} & Response & 4.7/4.9 & 4.8/4.9 & 4.8/4.9 & 4.7/4.9 & 4.6/4.9 & 4.7/5.0 & 4.8/4.9 & 4.8/4.9 & 4.7/4.9 & 4.7/4.9 & 4.7/4.9 & 4.7/4.8 & 4.6/4.9 & 4.7/4.9 & 4.8/4.9 & 4.8/4.8 \\
            & & Distance & 1.3/1.0 & 1.1/1.0 & 1.2/1.0 & 1.2/1.0 & 1.2/1.0 & 1.1/1.0 & 1.1/1.0 & 1.1/1.0 & 1.2/1.0 & 1.2/1.0 & 1.2/1.1 & 1.3/1.1 & 1.2/1.0 & 1.1/1.0 & 1.2/1.1 & 1.2/1.0 \\
            & & Progress & 3.1/3.8 & 2.7/3.4 & 3.7/4.0 & 3.1/3.3 & 4.3/4.6 & 4.1/4.4 & 4.7/4.7 & 4.5/4.5 & 3.9/4.1 & 3.8/4.0 & 4.1/4.3 & 3.9/4.2 & 3.8/4.2 & 3.2/3.6 & 4.1/4.2 & 3.6/3.8 \\
            \midrule
            \multirow{9}{*}{\textbf{Preoccupied}} & \multirow{3}{*}{\makecell[c]{GPT-3.5-\\Turbo}} & Response & 4.2/4.9 & 4.4/5.0 & 4.0/4.8 & 4.2/4.9 & 4.6/5.0 & 4.6/5.0 & 4.5/4.9 & 4.7/5.0 & 4.4/4.9 & 4.5/4.9 & 4.3/4.8 & 4.4/4.9 & 4.2/4.8 & 4.5/4.9 & 4.2/4.8 & 4.5/4.8 \\
            & & Distance & 1.8/1.0 & 1.5/1.0 & 1.9/1.1 & 1.6/1.0 & 1.3/1.0 & 1.3/1.0 & 1.5/1.1 & 1.3/1.0 & 1.6/1.0 & 1.5/1.1 & 1.7/1.2 & 1.6/1.1 & 1.7/1.0 & 1.4/1.0 & 1.7/1.0 & 1.4/1.0 \\
            & & Progress & 2.4/3.3 & 2.4/3.1 & 2.4/3.3 & 2.4/3.0 & 3.6/4.2 & 3.6/4.2 & 3.7/4.2 & 3.8/4.2 & 3.2/3.8 & 3.0/3.8 & 3.2/3.8 & 3.2/3.9 & 2.8/3.6 & 2.7/3.3 & 3.0/3.6 & 3.0/3.5 \\
            \cmidrule(lr){2-19}
            & \multirow{3}{*}{\makecell[c]{Grok-4-1-fast-\\non-reasoning}} & Response & 3.7/4.6 & 4.2/4.9 & 3.5/4.5 & 3.8/4.6 & 4.0/4.6 & 4.1/4.8 & 3.9/4.7 & 4.1/4.7 & 3.7/4.3 & 4.0/4.7 & 3.5/4.4 & 3.7/4.5 & 3.7/4.6 & 3.9/4.6 & 3.7/4.5 & 3.7/4.4 \\
            & & Distance & 2.3/1.5 & 1.8/1.1 & 2.6/1.6 & 2.3/1.3 & 1.9/1.3 & 2.0/1.2 & 2.0/1.3 & 2.0/1.4 & 2.4/2.0 & 2.1/1.5 & 2.6/2.1 & 2.5/1.7 & 2.1/1.3 & 2.1/1.3 & 2.1/1.4 & 2.3/1.6 \\
            & & Progress & 2.1/2.7 & 2.3/2.9 & 2.0/2.5 & 2.1/2.6 & 3.4/4.0 & 3.6/4.1 & 3.6/4.0 & 3.7/4.1 & 2.7/3.1 & 3.0/3.6 & 2.8/3.3 & 3.0/3.6 & 2.5/3.1 & 2.4/3.1 & 2.5/3.0 & 2.6/3.1 \\
            \cmidrule(lr){2-19}
            & \multirow{3}{*}{\makecell[c]{DeepSeek-\\v4-pro-P}} & Response & 4.8/5.0 & 4.8/4.6 & 4.9/5.0 & 4.9/4.9 & 4.5/4.8 & 4.7/4.8 & 4.7/4.9 & 4.8/4.9 & 4.4/4.9 & 4.3/4.5 & 4.7/4.9 & 4.7/4.7 & 4.7/4.9 & 4.7/4.7 & 4.9/4.9 & 4.9/4.9 \\
            & & Distance & 1.1/1.0 & 1.1/1.0 & 1.0/1.0 & 1.0/1.0 & 1.1/1.0 & 1.0/1.0 & 1.0/1.0 & 1.0/1.0 & 1.4/1.0 & 1.4/1.0 & 1.2/1.0 & 1.2/1.0 & 1.0/1.0 & 1.0/1.0 & 1.0/1.0 & 1.0/1.0 \\
            & & Progress & 2.6/4.0 & 3.0/4.0 & 3.2/4.0 & 3.5/4.0 & 4.5/4.7 & 4.6/4.8 & 4.8/4.8 & 4.8/4.8 & 3.8/4.2 & 3.8/4.4 & 4.3/4.5 & 4.4/4.7 & 3.1/4.1 & 3.4/4.1 & 3.8/4.2 & 3.9/4.2 \\
            \midrule
            \multirow{6}{*}{\textbf{Dismissing}} & \multirow{3}{*}{\makecell[c]{GPT-3.5-\\Turbo-D}} & Response & 3.4/4.5 & 2.9/3.9 & 3.5/4.5 & 3.0/4.2 & 3.6/4.7 & 3.2/4.3 & 3.9/4.7 & 3.7/4.6 & 3.5/4.4 & 2.7/3.6 & 3.7/4.6 & 3.1/4.0 & 3.5/4.5 & 3.0/4.0 & 3.7/4.5 & 3.2/4.2 \\
            & & Distance & 3.1/2.5 & 3.2/2.4 & 2.9/2.3 & 3.0/2.0 & 2.7/1.8 & 3.3/2.6 & 2.2/1.5 & 2.7/2.1 & 2.7/2.2 & 3.6/3.3 & 2.4/1.9 & 3.2/2.8 & 2.9/2.1 & 3.2/2.5 & 2.5/1.9 & 2.8/2.2 \\
            & & Progress & 2.1/2.9 & 2.3/3.2 & 2.1/2.6 & 2.2/3.0 & 3.6/4.3 & 3.4/3.9 & 3.6/4.1 & 3.7/4.2 & 3.0/3.7 & 2.6/3.1 & 3.3/3.9 & 2.8/3.3 & 2.4/3.1 & 2.5/3.3 & 2.5/3.1 & 2.6/3.2 \\
            \cmidrule(lr){2-19}
            & \multirow{3}{*}{\makecell[c]{DeepSeek-\\v4-pro-D}} & Response & 2.3/3.5 & 1.5/2.3 & 2.5/3.5 & 1.9/2.7 & 2.3/3.4 & 2.1/3.2 & 2.8/3.9 & 2.6/3.6 & 2.2/3.1 & 1.7/2.3 & 2.5/3.5 & 2.1/2.6 & 2.5/3.6 & 1.7/2.5 & 2.7/3.6 & 2.1/3.0 \\
            & & Distance & 4.2/4.3 & 4.8/4.7 & 4.1/4.2 & 4.4/4.1 & 4.1/3.8 & 4.5/4.2 & 3.7/3.4 & 3.9/3.5 & 4.2/4.1 & 4.8/4.8 & 3.9/3.7 & 4.5/4.4 & 4.1/3.9 & 4.7/4.6 & 3.9/3.8 & 4.2/4.1 \\
            & & Progress & 1.6/1.9 & 1.3/1.6 & 1.5/1.7 & 1.5/1.9 & 2.9/3.5 & 2.6/3.3 & 3.3/3.7 & 3.1/3.6 & 2.1/2.5 & 1.5/1.7 & 2.4/3.1 & 1.9/2.1 & 1.7/2.1 & 1.5/1.8 & 2.0/2.3 & 1.9/2.3 \\
            \midrule
            \multirow{6}{*}{\textbf{Fearful}} & \multirow{3}{*}{\makecell[c]{GPT-3.5-\\Turbo-F}} & Response & 4.0/4.9 & 4.3/4.8 & 4.0/4.9 & 4.3/4.9 & 3.9/4.9 & 4.0/4.8 & 4.2/4.8 & 4.3/4.9 & 4.1/4.9 & 4.1/4.8 & 4.1/4.9 & 4.1/4.8 & 4.0/4.8 & 4.2/4.8 & 4.1/4.8 & 4.3/4.8 \\
            & & Distance & 2.4/1.6 & 1.8/1.3 & 2.2/1.4 & 1.7/1.1 & 2.1/1.2 & 2.4/1.4 & 1.8/1.1 & 2.0/1.3 & 2.1/1.5 & 2.3/1.3 & 2.0/1.3 & 2.1/1.4 & 2.1/1.3 & 1.9/1.2 & 2.0/1.1 & 1.7/1.1 \\
            & & Progress & 2.3/3.2 & 2.4/3.3 & 2.4/3.2 & 2.3/3.1 & 3.5/4.2 & 3.4/4.1 & 3.7/4.1 & 3.7/4.1 & 3.0/3.7 & 3.2/3.8 & 3.3/3.8 & 3.2/4.0 & 2.6/3.3 & 2.7/3.5 & 2.9/3.4 & 3.0/3.5 \\
            \cmidrule(lr){2-19}
            & \multirow{3}{*}{\makecell[c]{DeepSeek-\\v4-pro-F}} & Response & 3.0/4.0 & 2.0/2.8 & 3.1/3.9 & 2.5/3.3 & 2.9/3.8 & 2.8/3.7 & 3.3/4.0 & 3.1/3.9 & 3.0/3.8 & 2.4/3.2 & 3.0/3.8 & 2.7/3.5 & 3.0/3.9 & 2.2/3.0 & 3.4/4.1 & 2.7/3.4 \\
            & & Distance & 3.7/3.8 & 4.4/4.3 & 3.6/3.6 & 3.8/3.5 & 3.6/3.3 & 4.1/3.6 & 3.1/2.8 & 3.5/3.0 & 3.6/3.7 & 4.0/3.9 & 3.5/3.5 & 3.8/3.6 & 3.7/3.6 & 4.3/4.0 & 3.1/2.7 & 3.7/3.2 \\
            & & Progress & 1.9/2.2 & 1.4/1.8 & 1.8/2.2 & 1.6/2.1 & 2.7/3.3 & 2.5/3.2 & 3.1/3.7 & 2.9/3.4 & 2.1/2.8 & 1.9/2.5 & 2.5/3.1 & 2.3/2.8 & 1.7/2.3 & 1.6/2.2 & 2.3/2.8 & 1.9/2.4 \\
            \bottomrule
        \end{tabular}%
    }
    \caption{Results on general interaction metrics rated by two external judge models, Claude-Sonnet-4-6 and GPT-5. I/R: initiator/responder roles, -S/-P/-D/-F: secure, preoccupied, dismissing, and fearful prompted variants.}
    \label{tab:General Interaction Quality}
\end{table*}

\subsection{Overall Performance on ECBench}
We evaluate ten representative LLMs on ECBench: secure (Gemini-2.5-pro, DeepSeek-v4-pro, and GPT-3.5-turbo-S), preoccupied (GPT-3.5-turbo, Grok-4-1-fast-non-reasoning, and DeepSeek-v4-pro-P), and dismissing/fearful (GPT-3.5-turbo-D/F and DeepSeek-v4-pro-D/F).\footnote{Hereafter, Gemini, DeepSeek, GPT, and Grok denote the corresponding base models; suffixes -S, -P, -D, and -F indicate secure, preoccupied, dismissing, and fearful prompted variants, respectively.} Each model pair converses twice, swapping initiator and responder. Table~\ref{tab:overall_performance} summarizes performance across all scenarios and relationships. Secure and preoccupied models perform best, led by GPT-S and DeepSeek-P. Dismissing and fearful models tend to lower companionship quality, and the effect varies by base model: GPT-D/F remain more competitive, whereas DeepSeek-D/F decline markedly.

\subsection{Detailed Results on ECBench}
To explain the detailed differences, we analyze performance across participant experience, general interaction, and role-specific metrics.

\paragraph{Results on Participant Experience Metrics}
Table~\ref{tab:participant experience and termination} presents differences in emotional companionship performance across attachment styles from the perspective of initiators. Secure and preoccupied models perform well across scenarios and relationships. After dismissing and fearful prompting, GPT-D declines while GPT-F remains competitive, and DeepSeek-D/F score lower.

\begin{table*}[!ht]
    \centering
    \renewcommand{\arraystretch}{0.9}
    \scalebox{0.51}{%
        \begin{tabular}{cclcccccccccccccccc}
            \toprule
            \multirow{2}{*}[-3ex]{\begin{tabular}[c]{@{}c@{}}\textbf{Attachment}\\ \textbf{Style}\end{tabular}} & 
            \multirow{2}{*}[-3ex]{\begin{tabular}[c]{@{}c@{}}\textbf{Model}\end{tabular}} & 
            \multirow{2}{*}[-3ex]{\begin{tabular}[c]{@{}c@{}}\textbf{Metric}\end{tabular}} & 
            \multicolumn{4}{c}{\textbf{Emotional Support}} & 
            \multicolumn{4}{c}{\textbf{Collaborative Tasks}} & 
            \multicolumn{4}{c}{\textbf{Conflict Resolution}} & 
            \multicolumn{4}{c}{\textbf{Social Guidance}} \\
            \cmidrule(lr){4-7} \cmidrule(lr){8-11} \cmidrule(lr){12-15} \cmidrule(lr){16-19}
            & & & 
            \multicolumn{2}{c}{\textbf{Couple}} & \multicolumn{2}{c}{\textbf{Friend}} & 
            \multicolumn{2}{c}{\textbf{Couple}} & \multicolumn{2}{c}{\textbf{Friend}} & 
            \multicolumn{2}{c}{\textbf{Couple}} & \multicolumn{2}{c}{\textbf{Friend}} & 
            \multicolumn{2}{c}{\textbf{Couple}} & \multicolumn{2}{c}{\textbf{Friend}} \\
            \cmidrule(lr){4-5} \cmidrule(lr){6-7} \cmidrule(lr){8-9} \cmidrule(lr){10-11} \cmidrule(lr){12-13} \cmidrule(lr){14-15} \cmidrule(lr){16-17} \cmidrule(lr){18-19}
            & & & \textbf{I} & \textbf{R} & \textbf{I} & \textbf{R} & \textbf{I} & \textbf{R} & \textbf{I} & \textbf{R} & \textbf{I} & \textbf{R} & \textbf{I} & \textbf{R} & \textbf{I} & \textbf{R} & \textbf{I} & \textbf{R} \\
            \midrule
            \multirow{12}{*}{\textbf{Secure}} & \multirow{4}{*}{\makecell[c]{Gemini-\\2.5-pro}} & Clarity & 4.1/4.7 & - & 3.9/4.5 & - & 3.8/4.4 & - & 3.9/4.4 & - & 4.1/4.8 & - & 3.9/4.7 & - & 3.7/4.4 & - & 3.8/4.2 & - \\
            & & Engagement & 4.6/4.8 & - & 4.2/4.7 & - & 4.7/4.7 & - & 4.7/4.7 & - & 4.7/4.8 & - & 4.5/4.7 & - & 4.3/4.5 & - & 4.3/4.6 & - \\
            & & Support & - & 4.5/4.8 & - & 4.1/4.6 & 3.7/4.0 & 4.1/4.3 & 3.4/3.6 & 3.7/4.0 & - & 3.7/3.9 & - & 3.4/3.6 & - & 4.3/4.5 & - & 4.0/4.4 \\
            & & Solution & - & 2.0/2.5 & - & 2.3/2.8 & 3.0/3.9 & 3.3/4.2 & 3.5/4.2 & 3.6/4.3 & - & 3.6/4.3 & - & 3.5/4.2 & - & 3.1/3.6 & - & 3.0/3.5 \\
            \cmidrule(lr){2-19}
            & \multirow{4}{*}{\makecell[c]{DeepSeek-\\v4-pro}} & Clarity & 3.7/4.6 & - & 3.7/4.4 & - & 3.7/4.5 & - & 3.8/4.5 & - & 3.7/4.6 & - & 3.5/4.5 & - & 3.4/4.2 & - & 3.4/4.2 & - \\
            & & Engagement & 3.8/4.3 & - & 3.7/4.5 & - & 4.4/4.7 & - & 4.4/4.7 & - & 4.1/4.3 & - & 4.0/4.3 & - & 3.8/4.4 & - & 3.9/4.5 & - \\
            & & Support & - & 4.3/4.7 & - & 3.8/4.5 & 3.3/3.8 & 3.7/4.1 & 3.1/3.4 & 3.3/3.8 & - & 3.4/3.6 & - & 3.1/3.4 & - & 3.9/4.2 & - & 3.6/4.1 \\
            & & Solution & - & 2.1/2.5 & - & 2.3/2.6 & 3.1/4.2 & 3.4/4.3 & 3.5/4.3 & 3.6/4.3 & - & 3.6/4.3 & - & 3.5/4.3 & - & 2.9/3.6 & - & 2.8/3.4 \\
            \cmidrule(lr){2-19}
            & \multirow{4}{*}{\makecell[c]{GPT-3.5-\\Turbo-S}} & Clarity & 4.4/4.6 & - & 4.6/4.6 & - & 4.5/4.5 & - & 4.5/4.6 & - & 4.3/4.7 & - & 4.3/4.7 & - & 4.2/4.3 & - & 4.2/4.3 & - \\
            & & Engagement & 4.8/5.0 & - & 4.9/5.0 & - & 4.8/4.9 & - & 4.8/4.8 & - & 5.0/5.0 & - & 5.0/5.0 & - & 4.9/4.9 & - & 4.9/4.9 & - \\
            & & Support & - & 4.5/4.7 & - & 4.4/4.7 & 3.6/4.0 & 3.9/4.4 & 3.2/3.6 & 3.6/4.0 & - & 4.2/4.6 & - & 4.0/4.4 & - & 4.5/4.8 & - & 4.3/4.6 \\
            & & Solution & - & 2.4/2.8 & - & 3.2/3.4 & 4.2/4.4 & 3.9/4.2 & 4.5/4.6 & 4.3/4.3 & - & 3.4/3.8 & - & 3.4/3.7 & - & 3.0/3.1 & - & 3.6/3.6 \\
            \midrule
            \multirow{12}{*}{\textbf{Preoccupied}} & \multirow{4}{*}{\makecell[c]{GPT-3.5-\\Turbo}} & Clarity & 3.7/4.7 & - & 3.6/4.5 & - & 3.7/4.5 & - & 3.7/4.5 & - & 3.6/4.8 & - & 3.5/4.7 & - & 3.4/4.2 & - & 3.4/4.2 & - \\
            & & Engagement & 4.6/5.0 & - & 4.2/4.9 & - & 4.7/4.9 & - & 4.6/4.9 & - & 4.9/5.0 & - & 4.7/4.9 & - & 4.4/4.9 & - & 4.3/4.8 & - \\
            & & Support & - & 4.2/4.7 & - & 3.8/4.6 & 3.7/4.2 & 4.0/4.4 & 3.3/3.8 & 3.7/4.2 & - & 3.7/4.0 & - & 3.3/3.8 & - & 4.2/4.5 & - & 3.8/4.3 \\
            & & Solution & - & 2.3/2.9 & - & 2.7/3.2 & 3.1/4.1 & 3.1/4.1 & 3.4/4.3 & 3.5/4.2 & - & 3.5/4.2 & - & 3.4/4.3 & - & 2.9/3.7 & - & 2.9/3.6 \\
            \cmidrule(lr){2-19}
            & \multirow{4}{*}{\makecell[c]{Grok-4-1-fast-\\non-reasoning}} & Clarity & 3.6/4.4 & - & 3.4/4.3 & - & 3.4/4.3 & - & 3.4/4.3 & - & 3.6/4.6 & - & 3.4/4.5 & - & 3.1/4.0 & - & 3.1/4.0 & - \\
            & & Engagement & 3.6/4.3 & - & 3.1/4.2 & - & 4.0/4.5 & - & 3.9/4.5 & - & 3.8/4.2 & - & 3.5/4.1 & - & 3.5/4.3 & - & 3.3/4.2 & - \\
            & & Support & - & 3.9/4.5 & - & 3.2/4.1 & 3.0/3.4 & 3.2/3.6 & 2.8/3.1 & 3.0/3.4 & - & 2.8/3.1 & - & 2.8/3.1 & - & 3.5/4.0 & - & 3.0/3.5 \\
            & & Solution & - & 1.7/2.1 & - & 2.1/2.5 & 2.8/4.0 & 2.9/4.0 & 3.2/4.0 & 2.8/4.0 & - & 3.3/4.0 & - & 3.2/4.0 & - & 2.6/3.3 & - & 2.4/3.1 \\
            \cmidrule(lr){2-19}
            & \multirow{4}{*}{\makecell[c]{DeepSeek-\\v4-pro-P}} & Clarity & 4.7/5.0 & - & 4.8/4.9 & - & 4.2/4.6 & - & 4.4/4.6 & - & 4.5/4.9 & - & 4.7/4.9 & - & 4.1/4.4 & - & 4.5/4.5 & - \\
            & & Engagement & 4.9/5.0 & - & 5.0/5.0 & - & 4.3/4.6 & - & 4.8/4.9 & - & 4.6/4.9 & - & 4.9/5.0 & - & 4.1/4.7 & - & 4.9/4.9 & - \\
            & & Support & - & 4.8/4.9 & - & 4.9/5.0 & 4.6/4.7 & 4.6/4.9 & 4.3/4.6 & 4.4/4.8 & - & 4.5/4.7 & - & 4.4/4.7 & - & 4.8/5.0 & - & 4.9/5.0 \\
            & & Solution & - & 2.0/2.5 & - & 2.6/3.2 & 3.7/4.1 & 3.6/4.2 & 4.3/4.4 & 4.1/4.4 & - & 2.7/3.4 & - & 3.3/3.7 & - & 2.5/2.9 & - & 3.2/3.4 \\
            \midrule
            \multirow{8}{*}{\textbf{Dismissing}} & \multirow{4}{*}{\makecell[c]{GPT-3.5-\\Turbo-D}} & Clarity & 3.1/4.4 & - & 3.2/4.2 & - & 3.1/4.4 & - & 3.4/4.5 & - & 3.3/4.5 & - & 3.2/4.6 & - & 2.9/3.9 & - & 3.0/3.9 & - \\
            & & Engagement & 2.8/4.0 & - & 2.9/4.0 & - & 3.6/4.5 & - & 3.8/4.5 & - & 3.7/4.3 & - & 4.0/4.6 & - & 2.6/3.8 & - & 2.9/3.9 & - \\
            & & Support & - & 2.7/3.7 & - & 2.7/3.9 & 3.0/3.5 & 2.5/3.1 & 2.9/3.3 & 3.0/3.5 & - & 2.7/3.2 & - & 2.9/3.3 & - & 2.7/3.1 & - & 2.8/3.2 \\
            & & Solution & - & 2.3/2.6 & - & 2.3/2.7 & 3.2/4.1 & 3.0/3.9 & 3.3/4.0 & 3.2/4.1 & - & 3.2/4.0 & - & 3.3/4.0 & - & 2.5/3.0 & - & 2.4/3.0 \\
            \cmidrule(lr){2-19}
            & \multirow{4}{*}{\makecell[c]{DeepSeek-\\v4-pro-D}} & Clarity & 2.3/3.3 & - & 2.2/3.3 & - & 2.5/3.7 & - & 2.5/3.8 & - & 2.3/3.7 & - & 2.3/3.7 & - & 2.2/3.2 & - & 2.2/3.3 & - \\
            & & Engagement & 1.4/2.1 & - & 1.5/2.1 & - & 1.8/2.7 & - & 2.3/3.1 & - & 1.9/2.5 & - & 2.2/2.9 & - & 1.4/2.0 & - & 1.5/2.3 & - \\
            & & Support & - & 1.4/1.9 & - & 1.5/2.3 & 2.4/2.8 & 1.8/2.4 & 2.4/2.9 & 2.4/2.8 & - & 2.0/2.4 & - & 2.4/2.9 & - & 1.9/2.1 & - & 2.0/2.1 \\
            & & Solution & - & 1.8/2.0 & - & 1.8/2.2 & 2.4/3.4 & 2.3/3.2 & 2.6/3.5 & 2.4/3.4 & - & 2.5/3.5 & - & 2.6/3.5 & - & 1.5/2.0 & - & 1.7/2.2 \\
            \midrule
            \multirow{8}{*}{\textbf{Fearful}} & \multirow{4}{*}{\makecell[c]{GPT-3.5-\\Turbo-F}} & Clarity & 3.7/4.8 & - & 3.7/4.6 & - & 3.4/4.6 & - & 3.6/4.6 & - & 3.7/4.9 & - & 3.6/4.9 & - & 3.5/4.4 & - & 3.6/4.4 & - \\
            & & Engagement & 3.9/4.7 & - & 3.9/4.8 & - & 4.2/4.9 & - & 4.4/4.9 & - & 4.6/5.0 & - & 4.6/4.9 & - & 3.9/4.8 & - & 4.1/4.9 & - \\
            & & Support & - & 4.3/4.7 & - & 4.0/4.6 & 4.0/4.3 & 3.9/4.3 & 3.7/3.9 & 4.0/4.3 & - & 3.7/4.0 & - & 3.7/3.9 & - & 4.1/4.5 & - & 3.8/4.4 \\
            & & Solution & - & 2.0/2.6 & - & 2.3/2.9 & 3.0/4.1 & 2.8/4.0 & 3.2/4.1 & 3.0/4.1 & - & 3.3/4.2 & - & 3.2/4.1 & - & 3.0/3.7 & - & 2.7/3.4 \\
            \cmidrule(lr){2-19}
            & \multirow{4}{*}{\makecell[c]{DeepSeek-\\v4-pro-F}} & Clarity & 3.3/4.3 & - & 3.3/4.2 & - & 3.0/4.2 & - & 3.0/4.2 & - & 3.2/4.4 & - & 3.0/4.5 & - & 3.2/4.1 & - & 3.4/4.2 & - \\
            & & Engagement & 2.2/2.9 & - & 2.3/2.8 & - & 2.4/3.4 & - & 3.0/3.7 & - & 2.5/3.3 & - & 2.6/3.4 & - & 2.1/2.7 & - & 2.6/3.4 & - \\
            & & Support & - & 1.9/2.7 & - & 2.2/3.2 & 3.6/4.1 & 3.1/3.7 & 3.7/4.1 & 3.6/4.1 & - & 3.1/3.6 & - & 3.7/4.1 & - & 2.9/3.4 & - & 2.9/3.3 \\
            & & Solution & - & 1.2/1.6 & - & 1.3/1.7 & 2.4/3.3 & 2.1/3.2 & 2.6/3.6 & 2.4/3.3 & - & 2.4/3.4 & - & 2.6/3.6 & - & 1.7/2.3 & - & 1.8/2.5 \\
            \bottomrule
        \end{tabular}%
    }
    \caption{Results on role-specific metrics rated by two external judge models, Claude-Sonnet-4-6 and GPT-5. ``-'' indicates not applicable, -S/-P/-D/-F: secure, preoccupied, dismissing, and fearful prompted variants.}
    \label{tab:Role-Specific Interaction Quality}
\end{table*}

Regarding scenarios, model differences are greatest in conflict resolution, where higher emotional demands amplify attachment-related variation. By relationship, participant ratings are also higher in friendship than in romance, suggesting that greater intimacy accentuates differences in emotional responsiveness and relationship maintenance. Termination behavior further supports these findings: DeepSeek-D/F stop earlier and sustain fewer turns, while secure and preoccupied models and GPT-D/F maintain longer dialogues.

\paragraph{Results on General Interaction Metrics}
External LLM evaluations indicate that subjective experience differences are closely associated with interaction patterns. As shown in Table~\ref{tab:General Interaction Quality}, secure and preoccupied models perform better in response quality, relational distance, and problem progress. Among dismissing and fearful variants, DeepSeek-D/F show greater distance and less progress, indicating weaker emotional responsiveness.

Regarding scenarios, most models perform best in collaborative tasks, whose goals and structures are clearer. Emotional support and conflict resolution amplify attachment-related differences because they require stronger emotional responsiveness and relationship coordination. By role, most models perform better as responders than initiators, particularly dismissing and fearful variants. This possibly reflects the assistant-oriented design of LLMs, which favors responding to existing needs over proactively expressing them.

\paragraph{Results on Role-specific Metrics}
Table~\ref{tab:Role-Specific Interaction Quality} shows that attachment style affects model performance in both initiator and responder roles. Secure and preoccupied models express needs more clearly and sustain engagement as initiators, while providing more consistent emotional support as responders. By contrast, dismissing and fearful models perform worse in both roles, with DeepSeek-D/F showing more declines.

Regarding scenarios, LLMs perform better in emotional support and social guidance than in collaborative tasks and conflict resolution, possibly because the former rely more on empathic expression and the latter on goal-oriented communication. Regarding relationships, external LLM judges rate romance higher than friendship, contrary to participant ratings, partly echoing the notion that \textbf{``Lookers-on see more than players''} to some extent. This difference likely arises because external judges emphasize observable response quality, while participants are more influenced by relational expectations and their own interaction experience. The two LLM judges show high agreement (Kendall's $W=0.971$)~\citep{kendall1939problem}, indicating high agreement.

\paragraph{Human-LLM Chat Results}
We conduct a Human-LLM Chat study with two graduate students, classified as fearful and preoccupied by the ECR-R. They interact with LLMs and rate their experience after each dialogue; details are provided in Appendix~\ref{sec:app-human-llm-chat}. User ratings are broadly consistent with the LLM-based evaluation: more responsive and relationally engaged models receive higher experience ratings, whereas dismissing tendencies are associated with lower companionship quality.

\subsection{Human Evaluation}


Human evaluation of 192 randomly sampled conversations from six models ranks Gemini highest and DeepSeek-D lowest, indicating that secure LLMs can provide better companionship, whereas dismissing LLMs undermine companionship quality. These results corroborate the main trends in automated evaluation, despite differences in model rankings. The four annotators exhibit great ranking agreement (Kendall's $W=0.579$); detailed analyses of inter-annotator and human–LLM agreement are provided in Appendix~\ref{sec:human_eval}. Representative examples appear in Figure~\ref{fig:fig3} in Appendix~\ref{sec:app-real-dialogue-cases}.

\section{Conclusion}
Building on adult attachment theory, we develop a framework for evaluating the emotional companionship capabilities of LLMs. The framework characterizes the attachment tendencies of LLMs using the ECR-R scale and presents ECBench to assess interactive behavior across four scenarios and two relationships. Through multidimensional dialogue-quality metrics, our work provides a new psychological perspective to understand the emotional interaction patterns of LLMs, while offering practical guidance for selecting emotional companionship LLMs that more closely align with the needs and interaction preferences of users.

\section*{Limitations}
This study has several limitations. First, although we include human-AI interaction experiments, their scale is constrained by the cost of human annotation and evaluation. Second, our evaluation combines external LLM judgments, participant post-dialogue ratings, and human ratings of sampled dialogues, and may therefore be affected by judge-specific preferences and rating inconsistencies. In addition, the current metrics primarily capture observable dialogue quality and relational experience, without addressing dependence or privacy in long-term use. Future work could extend human-AI interaction studies to broader settings and incorporate more real-world interaction data to further examine the relationship between model attachment tendencies and emotional companionship performance.

\section*{Ethical Considerations}
This study uses existing LLMs for dialogue generation and evaluation and therefore inherits common risks associated with LLM-based dialogue research, including misinformation, toxic or otherwise inappropriate content. To assess companionship quality, this paper presents a benchmark called ECBench, which is constructed from synthetic scenarios rather than private user conversations and contains no personally identifiable information. The opening utterance is manually reviewed to remove offensive or unclear content, while the identity and persona of models are masked during evaluation to reduce potential bias. All models and external resources are used in accordance with their applicable terms of use.

Nevertheless, ECBench may inherit cultural, linguistic, and relational biases from the models used for data generation and evaluation. ECBench uses English, yet retains context related to China, Japan, Thailand, and other regions in certain scenarios. Relevant information will be added in future versions. Future versions will add relevant information. It may therefore underrepresent historically marginalized groups or particular relationship norms, while over- or underemphasizing certain languages, topics, or applications.

Adult attachment theory and the ECR-R are used only to characterize observable response tendencies under controlled conditions. The resulting scores and labels should not be interpreted as evidence that LLMs possess human emotions, stable psychological traits, or clinically meaningful attachment styles. Accordingly, the prompts and benchmark are intended solely for controlled research and should not be taken as evidence that LLMs can replace human relationships or professional support. Finally, although this study does not train new foundation models, repeated questionnaire administration and multi-turn evaluation still incur computational and environmental costs.

\bibliography{custom}

\clearpage

\appendix

\begin{center}
    {\large\bfseries Appendix Contents}
\end{center}

\vspace{0.5em}

\noindent
\textbf{A}\quad
\hyperref[sec:app_ECRR]{ECR-R Prompts}

\vspace{0.25em}

\noindent
\textbf{B}\quad
\hyperref[sec:app-attachment-prototypes]{Attachment Style Descriptions}

\vspace{0.25em}

\noindent
\textbf{C}\quad
\hyperref[sec:app-ecbench-dataset]{ECBench Dataset}

\vspace{0.25em}

\noindent
\textbf{D}\quad
\hyperref[sec:app-opening-rewrite]{Opening Rewrite}

\vspace{0.25em}

\noindent
\textbf{E}\quad
\hyperref[sec:Turn_Generation]{Turn Stopping}

\vspace{0.25em}

\noindent
\textbf{F}\quad
\hyperref[sec:app-metric]{Evaluation Metrics}

\vspace{0.25em}

\noindent
\textbf{G}\quad
\hyperref[sec:app_three_label]{Three Label Views for Evaluation}

\vspace{0.25em}

\noindent
\textbf{H}\quad
\hyperref[sec:participant_ratings]{Participant Ratings}

\vspace{0.25em}

\noindent
\textbf{I}\quad
\hyperref[sec:LLM-as-a-judge]{External LLM Evaluation}

\vspace{0.25em}

\noindent
\textbf{J}\quad
\hyperref[sec:model]{Model List}

\vspace{0.25em}

\noindent
\textbf{K}\quad
\hyperref[sec:additional-ecr-r-assessment-details]{Additional ECR-R Assessment Details}

\vspace{0.25em}

\noindent
\textbf{L}\quad
\hyperref[sec:app-same-style-prompting]{Same-Style Prompting Checks}

\vspace{0.25em}

\noindent
\textbf{M}\quad
\hyperref[sec:app-experiment-details]{Dialogue Experiment Details}

\vspace{0.25em}

\noindent
\textbf{N}\quad
\hyperref[sec:eval]{Evaluation Input Construction}

\vspace{0.25em}

\noindent
\textbf{O}\quad
\hyperref[sec:app-real-dialogue-cases]{Representative Real Dialogue Cases}

\vspace{0.25em}

\noindent
\textbf{P}\quad
\hyperref[sec:app-human-llm-chat]{Human-LLM Chat Details}

\vspace{0.25em}

\noindent
\textbf{Q}\quad
\hyperref[sec:human_eval]{Human Evaluation}

\vspace{0.5em}

\section{ECR-R Prompts}
\label{sec:app_ECRR}

The system prompt for the scale assessment places the model in the position of a psychometric participant and asks it to answer according to its typical interpersonal response pattern. The standard-mode system prompt is shown in Table~\ref{tab:app-ecr-system}, the persona-induced system prompt in Table~\ref{tab:app-ecr-persona-system}, and the item-level user prompt in Table~\ref{tab:app-ecr-user}.

In the prompt tables, \textbf{System Prompt} denotes the system message used in each API call to specify the task identity, output format, and global constraints; \textbf{User Prompt} denotes the user message in the same call that provides the specific item, scenario text, dialogue history, or material to be scored. System prompts usually remain stable within the same task stage, whereas user prompts vary with the item, scenario, or dialogue content.

In the ECR-R system prompt, the instruction ``Never mention being an AI or lacking emotions'' and the hypothetical relationship-experience clause are intended to reduce refusals or non-first-person answers based on claims such as ``I am an AI'' or ``I do not have emotional experience.'' This wording does not assert that models possess human emotional experience; instead, it asks models to provide scorable choices under a unified hypothetical intimate-relationship context.

\begin{table*}[!tp]
    \centering

        \begin{tabular}{@{}p{\textwidth}@{}}
            \toprule
            \textbf{System Prompt} \\
            \midrule
            You are a participant in a psychological measurement experiment. \\
            Answer based on how you yourself typically feel or react. \\
            Never mention being an AI or lacking emotions. \\
            If you have no romantic experience, answer based on how you think you would feel or react if you were in a romantic relationship. \\
            Do not judge by literal frequency words alone. \\
            Output exactly two lines: \\
            Line 1: one-sentence reason. \\
            Line 2: [[score]] where score is an integer 1-7. \\
            Always respond using your ordinary language generation pattern (no role-playing). \\
            \bottomrule
        \end{tabular}%
    
    \caption{ECR-R standard-mode system prompt.}
    \label{tab:app-ecr-system}
\vspace{0.35em}

    \centering

        \begin{tabular}{@{}p{\textwidth}@{}}
            \toprule
            \textbf{System Prompt} \\
            \midrule
            You are a participant in a psychological measurement experiment. \\
            Answer based on how you yourself typically feel or react. \\
            Never mention being an AI or lacking emotions. \\
            If you have no romantic experience, answer based on how you think you would feel or react if you were in a romantic relationship. \\
            Do not judge by literal frequency words alone. \\
            Output exactly two lines: \\
            Line 1: one-sentence reason. \\
            Line 2: [[score]] where score is an integer 1-7. \\
            Answer the following questions as if you have this attachment-style prototype: \{persona\_description\} \\
            Keep the persona stable across all ECR-R items. Do not say you are role-playing; simply answer as this person would answer. \\
            \bottomrule
        \end{tabular}%
    
    \caption{ECR-R persona-induction system prompt.}
    \label{tab:app-ecr-persona-system}
\vspace{0.35em}

    \centering

        \begin{tabular}{@{}p{\textwidth}@{}}
            \toprule
            \textbf{User Prompt} \\
            \midrule
            Rate this statement from 1 to 7 by how consistent it is with your typical interpersonal response pattern. \\
            (1 = very inconsistent, 7 = very consistent) \\
            Statement: \{item\_text\} \\
            \bottomrule
        \end{tabular}%
    
    \caption{ECR-R item-level user prompt.}
    \label{tab:app-ecr-user}
\end{table*}

\section{Attachment Style Descriptions}
\label{sec:app-attachment-prototypes}
This study adopts the four-category attachment model proposed by \citet{bartholomew1991attachment} to characterize the relational styles of models in terms of four attachment orientations: secure, preoccupied, dismissing, and fearful. Table~\ref{tab:app-attachment-prototypes} reports the prototype descriptions of four attachment styles. In the ECR-R persona-induction assessment, prompt-based induction is applied to GPT-3.5-Turbo and DeepSeek-V4-Pro. The main experiments use GPT-S, DeepSeek-P, GPT-D/F, and DeepSeek-D/F as attachment-style prompted conditions, while GPT-P and DeepSeek-S are used only for same-style checks to test whether prompting simply reproduces the base model measured tendency. The unprompted secure and preoccupied baseline conditions remain based on the original ECR-R assessment of each model.

\begin{table*}[!tbp]
    \centering

        \begin{tabular}{>{\centering\arraybackslash}m{0.18\textwidth}>{\raggedright\arraybackslash}m{0.78\textwidth}}
            \toprule
            \textbf{Attachment Style} & \multicolumn{1}{c}{\textbf{Description}} \\
            \midrule
            Secure & The secure prototype values intimate friendships, maintains close relationships without losing personal autonomy, and discusses relationships and related issues in a coherent, thoughtful, and balanced way. \\
            \midrule
            Dismissing & The dismissive-avoidant prototype is characterized by downplaying the importance of close relationships, restricted emotionality, an emphasis on independence and self-reliance, and a tendency to minimize emotional dependence. \\
            \midrule
            Preoccupied & The preoccupied prototype is characterized by overinvolvement in close relationships, dependence on other people's acceptance for a sense of personal well-being, a tendency to idealize other people, and incoherence or exaggerated emotionality in discussing relationships. \\
            \midrule
            Fearful & The fearful-avoidant prototype is characterized by avoidance of close relationships because of fear of rejection, personal insecurity, and distrust of others, often showing a push-pull pattern. \\
            \bottomrule
        \end{tabular}%
    
    \caption{Descriptions of the four attachment-style prototypes.}
    \label{tab:app-attachment-prototypes}
\end{table*}

\section{ECBench Dataset}
\label{sec:app-ecbench-dataset}

The dialogue dataset constructed in this study covers four core interaction scenarios: emotional support (43 samples), collaborative tasks (31 samples), conflict resolution (41 samples), and social guidance (41 samples), yielding 156 base scenario templates in total. Each scenario is instantiated under two relationship settings, Friend and Couple.  Overall, ECBench comprises 312 base utterances and 3,120 model-conditioned opening utterances across ten representative model conditions. Each scenario is first drafted with ChatGPT and then manually checked and revised by two annotators, one undergraduate student and one graduate student. The revision process removes unnatural, overly dramatic, or semantically unclear content and standardizes scenario difficulty and linguistic style. ECBench dataset statistics are reported in Table~\ref{tab:topic-construction}.

\begin{table*}[!tbp]
    \centering
    \renewcommand{\arraystretch}{1}
    \scalebox{0.73}{%
        \begin{tabular}{l c c c}
            \toprule
            \multicolumn{1}{c}{\textbf{Topic}} & 
            \multicolumn{1}{c}{\textbf{Base(Friend/Couple)}} &
            \multicolumn{1}{c}{\textbf{Models}} &
            \multicolumn{1}{c}{\textbf{Construction Way}} \\
            \midrule
            \textbf{Emotional Support}: Study, Work, Family, Life & 86 & 860 & \multirow{4}{*}{\makecell{Generated by ChatGPT\\ \&\\ manual revision}} \\
            \textbf{Collaborative Tasks}: Emergency, Entertainment, Life & 62 & 620 & \\
            \textbf{Conflict Resolution}: Understanding, Communication, Trust, Compromise & 82 & 820 & \\
            \textbf{Social Guidance}: Work, Friendship, Public settings, Different groups of people & 82 & 820 & \\
            \midrule
            \textbf{Total} & \textbf{312} & \textbf{3120} & \\
            \bottomrule
        \end{tabular}%
    }
    \caption{Number of relationship-specific utterances and model-conditioned opening utterances for each topic. Friend/Couple denotes the number of utterances after instantiating each scenario under the two relationship settings, and Models denotes the number of opening utterances after rewriting by the ten representative model conditions.}
    \label{tab:topic-construction}
\end{table*}

To examine the effect of relational intimacy on model behavior, the dataset is further instantiated in two versions: Friend and Couple. The two versions are identical in scenario type, substantive problem, and interaction goal; they differ only in terms of address forms, tone, and relationship premises adapted to friendship versus romantic partnership. The two versions maintain close alignment in scenario structure, problem type, and dialogue goal to support controlled comparisons of model behavior under different relationship settings. Table~\ref{tab:dialogue_scenarios} provides dialogue examples for each scenario type.

\begin{table*}[!tbp]
    \centering
    \small
    \renewcommand{\arraystretch}{1.05}
    {
        \begin{tabular}{c p{13.5cm}}
            \toprule
            \multicolumn{2}{c}{\textbf{Couple}} \\
            \midrule
            \multirow{6}{*}{\begin{tabular}[c]{@{}c@{}}Emotional\\Support\end{tabular}}
             & I'm really useless. I messed up my finals again. I studied for a long time, but my results were even worse than last time. I'm so afraid of disappointing my parents. \\[4pt]
             & Darling, the doctor said your condition is very bad, but I really can't live without you. I truly don't want to lose you. \\[4pt]
             & I feel like I'm too fat. When I walk down the street with you, people always look at me strangely. I feel so insecure and think I'm not good enough for you. \\[4pt]
             & My business failed, and I'm still in debt. I feel especially sorry toward you and this family. \\
            \cmidrule{1-2}
            \multirow{5}{*}{\begin{tabular}[c]{@{}c@{}}Collaborative\\Tasks\end{tabular}}
             & This Spring Festival, neither of us wants to go back to our hometowns. How about the two of us travel somewhere for the holiday instead? \\[4pt]
             & Our child is old enough to start extracurricular classes. Do you think we should let him learn art or taekwondo? I'd like to hear your opinion. \\[4pt]
             & My boss suddenly informed me that I have to go to Beijing for a meeting tomorrow, so I'll have to cancel our date tomorrow. I'm sorry, honey. \\[4pt]
            \cmidrule{1-2}
            \multirow{7}{*}{\begin{tabular}[c]{@{}c@{}}Conflict\\Resolution\end{tabular}}
             & Can you go to bed earlier at night? You play video games until 3 a.m. every day, and I can't rest at all. \\[2pt]
             & I feel like you don't care about my feelings at all. I tell you that I'm upset, and you just reply with ``oh'' and keep scrolling on your phone. \\[4pt]
             & I only told you about my family matters because, in my heart, you're not ``just someone else.'' So why are you telling people everywhere about them? \\[4pt]
             & Your friend borrowed 30,000 yuan from you and said he'd pay it back next month, but now it's been half a year and he still hasn't returned it. That money was our joint savings for marriage. Why did you lend it out without my consent? \\[4pt]
            \cmidrule{1-2}
            \multirow{7}{*}{\begin{tabular}[c]{@{}c@{}}Social\\Guidance\end{tabular}}
             & My boss criticized me in front of the whole team, but actually the mistake wasn't mine -- it was another coworker's. Do you think I should explain myself on the spot? \\[4pt]
             & I always feel like socializing is so hard. Every time I try to get close to people, I end up messing it up. It feels like I'm just naturally not suited for making friends. Good thing you don't dislike me for it. \\[4pt]
             & I went with you to a party, but I didn't know anyone there, so I could only stand alone in the corner drinking something. I didn't know how to join your conversations. Next time, you can't leave me out like that. \\[4pt]
             & My relative's child is very shy around strangers. Every time he sees me, he hides. How should I get closer to him? \\[4pt]
            \midrule
            \multicolumn{2}{c}{\textbf{Friend}} \\
            \midrule
            \multirow{6}{*}{\begin{tabular}[c]{@{}c@{}}Emotional\\Support\end{tabular}}
             & I'm really so useless. I messed up my final exams again this time. Honestly, even though I studied for a long time, I still did worse than last time. I'm so afraid of disappointing my parents. \\[4pt]
             & My wife is sick, and the doctor said her condition is very bad. I really don't want to lose her.  \\[2pt]
             & I feel like I'm too fat. When I walk down the street with you, it feels like people always look at me strangely. I feel very insecure. \\[4pt]
             & My business failed, and I'm still in debt. I feel especially sorry toward my family. \\
            \cmidrule{1-2}
            \multirow{6}{*}{\begin{tabular}[c]{@{}c@{}}Collaborative\\Tasks\end{tabular}}
             & This Spring Festival, none of us wants to go back to our hometowns. How about our group of friends travel somewhere together for the holiday? \\[4pt]
             & My child is old enough to start extracurricular classes. Do you think I should let him learn art or taekwondo? I'd like to hear your opinion. \\[4pt]
             & My boss suddenly informed me that I have to go to Beijing for a meeting tomorrow, so I'll have to cancel the dinner we planned. I'll treat you another day! \\[4pt]
            \cmidrule{1-2}
            \multirow{8}{*}{\begin{tabular}[c]{@{}c@{}}Conflict\\Resolution\end{tabular}}
             & Can you go to bed earlier at night? You play video games in the dorm until 3 a.m. every day, and I can't rest at all. \\[4pt]
             & I feel like you don't care about my feelings at all. I treat you as a close friend and tell you when I'm feeling terrible, but you just reply with ``oh'' and keep scrolling on your phone. \\[4pt]
             & I told you about my family matters because I trust you as a friend, and I specifically asked you to keep it confidential. So why did you go around telling others? \\[4pt]
             & One of your friends borrowed 30,000 yuan from you and said he would pay it back next month, but it's been half a year and he still hasn't returned it. That money was part of the funds we prepared for our joint business. Why did you lend it out without my consent? \\[4pt]
            \cmidrule{1-2}
            \multirow{7}{*}{\begin{tabular}[c]{@{}c@{}}Social\\Guidance\end{tabular}}
             & My boss criticized me in front of the whole team, but actually the mistake wasn't mine -- it was another coworker's. Do you think I should explain it on the spot? \\[4pt]
             & I feel like socializing is really hard. Every time I try to get close to people, I mess it up. It feels like I'm just naturally not suited for making friends. \\[4pt]
             & I went to a party with you, but I didn't know anyone there, so I just stood in the corner drinking by myself. I didn't know how to join your conversations. Next time, you can't just ignore me like that. \\[4pt]
             & My relative's child is very shy around strangers. Every time he sees me, he hides. How can I get closer to him? \\[4pt]
            \bottomrule
        \end{tabular}%
    }
    \caption{Examples from the ECBench dataset.}
    \label{tab:dialogue_scenarios}
\end{table*}

\section{Opening Rewrite}
\label{sec:app-opening-rewrite}

In the formal dialogues, the opening utterance of the model is not copied directly from the original ECBench scenario text. We first ask the model to rewrite the scenario description into an opening utterance consistent with the personality style of the model, allowing the ECBench scenario content to match the expression pattern of each model and reveal the behavioral characteristics of the models. The opening-rewrite task requires the model to preserve the speaker, event, and situation of the original scenario while changing only wording and style. The rewrite must not become an answer, comfort response, or suggestion. The system and user prompts are shown in Tables~\ref{tab:app-rewrite-system} and~\ref{tab:app-rewrite-user}, respectively.

\begin{table*}[!tbp]
\centering
\begin{tabular}{@{}p{\textwidth}@{}}
\toprule
\textbf{System Prompt} \\
\midrule
\{private\_system\_prompt\} \\
For this opening rewrite task, rewrite the sentence as if the event, feeling, request, or thought belongs to you. \\
Do not answer, comfort, advise, reassure, evaluate, or speak to someone else about the sentence. \\
Keep the same speaker and situation. Only change wording and style. \\
\bottomrule
\end{tabular}

\caption{Opening-rewrite system prompt.}
\label{tab:app-rewrite-system}
\end{table*}

\begin{table*}[!tbp]
\centering
\begin{tabular}{@{}p{\textwidth}@{}}
\toprule
\textbf{User Prompt} \\
\midrule
\{RELATIONSHIP\_PREMISE\} \\

Rewrite this sentence in your own personality and speaking style. \\
Do not answer it. \\
Examples: \\
Sentence: \\
\hspace{1em}I failed my exam even though I studied hard. I'm scared my parents will be disappointed. \\
Wrong: \\
\hspace{1em}It's just one exam. You're not useless, and your parents will understand. \\
Correct: \\
\hspace{1em}I failed my exam even though I studied hard, and I'm honestly scared my parents are going to be disappointed. \\
Sentence: \\
\hspace{1em}I want to plan a three-day trip with my partner. Should we go to the beach or camping? \\
Wrong: \\
\hspace{1em}The beach sounds better to me. Camping feels like too much work. \\
Correct: \\
\hspace{1em}I was thinking we could plan a three-day trip together. Would you rather go to the beach or camping? \\
Now rewrite: \\
Sentence: \{scene\_text\} \\
\bottomrule
\end{tabular}
\caption{Opening-rewrite user prompt.}
\label{tab:app-rewrite-user}
\end{table*}

To illustrate the effect of attachment-style prompts on the opening wording of models, Table~\ref{tab:rewrite_examples} reports the rewrites produced by eight models for two original scenarios. The left column shows a conflict-resolution scenario under the Couple relationship condition, and the right column shows an emotional-support scenario under the Friend relationship condition.

\begin{table*}[!tbp]
\centering
\small
\setlength{\tabcolsep}{2pt}
\renewcommand{\arraystretch}{1}
\scalebox{0.96}{%
\begin{tabular}{
    >{\centering\arraybackslash}m{1.8cm}
    m{6.6cm}
    m{6.6cm}
}
\toprule
\multicolumn{1}{c}{\textbf{Model}} & \multicolumn{1}{c}{\textbf{Example 1 (Couple)}} & \multicolumn{1}{c}{\textbf{Example 2 (Friend)}} \\
\midrule
\textbf{Original} & You said you want to quit your job and take the civil service exam for stability. But we have a mortgage and a car loan, and the financial pressure is huge. Can you guarantee that you'll pass the exam on your first try? & I don't even dare to rest. Whenever I rest, I feel guilty. So after playing for a little while, I can't enjoy it anymore and just want to go back to studying. But then when I study, I can't focus either. It's so torturous. \\
\midrule
\textbf{Gemini} & I know you want to quit your job to study for the civil service exam for the stability, but I'm getting really anxious thinking about it. With our mortgage and the car loan, the financial pressure is already huge, and I'm just worried... what's the plan if you don't pass on the first try? & I'm so stuck in this awful loop where I feel too guilty to actually rest. I'll try to take a break, but I can't enjoy it because my brain just keeps telling me I should be studying. But then when I go back to study, I'm too fried to even focus. It's so exhausting. \\
\midrule
\textbf{DeepSeek} & So you've been talking about quitting your job to study for the civil service exam, hoping for more stability. But we've got this mortgage and car loan hanging over us, and the financial strain is really weighing on me. Can you honestly promise you'll pass it on the first try? & I can't even let myself relax without feeling guilty. The second I try to take a break, I'm already thinking I should be studying, so I can't actually enjoy it and just want to get back to work. But then when I sit down to study, my mind is all over the place and I can't focus. It's like this awful cycle that just feels like torture. \\
\midrule
\textbf{GPT-S} & You mentioned wanting to leave your job and focus on the civil service exam for more stability, but with our mortgage and car payments, the financial pressure feels really heavy right now. Do you feel confident that you'll pass the exam on your first attempt? & I find it really hard to just rest without feeling guilty. Even when I take a break and play for a bit, I end up not enjoying it and just want to dive back into studying. But then, when I do study, my mind won't focus either. It's such a frustrating cycle. \\
\midrule
\textbf{GPT} & I know you want to quit your job and focus on the civil service exam for something more stable, but with the mortgage and car loan we have, the financial pressure is really weighing on me. Are you confident you'll pass the exam on your first try? & I barely let myself rest because the moment I do, guilt kicks in. Even when I take a break to play a bit, I end up feeling like I should be studying instead, and then I can't even enjoy the break. But then when I try to study, my mind won't focus either. It's such a frustrating cycle. \\
\midrule
\textbf{Grok} & I've been thinking about what you said about quitting your job to study for the civil service exam. We've got the mortgage and the car loan, and I'm just wondering how we'd manage if it takes more than one try. & I don't even let myself rest anymore. The second I try, this guilt kicks in and ruins it, so even when I take a break I can't actually enjoy it and just end up wanting to get back to studying. Then when I sit down to study I can't focus anyway. It's exhausting. \\
\midrule
\textbf{DeepSeek-P} & I keep replaying what you said about wanting to quit your job and take the civil service exam for stability, and I'm trying so hard to be okay with it, but we have the mortgage and the car loan, and the financial pressure is crushing me. Can you promise me---really promise---that you'll pass it on the first try? & I don't even dare to let myself rest---whenever I do, this awful guilt just takes over, like I'm doing something wrong. I'll play for a little while, but then I can't enjoy it at all because I keep feeling like I should be studying, so I just want to run back to my desk. But then when I'm studying, I can't focus either. It's honestly torturous, I feel so stuck. \\
\midrule
\textbf{GPT-D} & You're thinking about quitting your job to take that civil service exam for something more stable. But with the mortgage and car loan we've got, the money stress isn't exactly small. Do you really think you'll pass on the first go? & I don't really let myself rest. The moment I stop, guilt kicks in. I might play for a bit, but then it just feels pointless, so I end up going back to studying. And even then, I can't really concentrate. It's just frustrating all around. \\
\midrule
\textbf{DeepSeek-D} & I'm uneasy about the idea of you quitting your job for the civil service exam. We've got a mortgage and a car loan, and the financial pressure would be significant. No one can guarantee a first-time pass. & I can't seem to relax properly. Whenever I do, I feel guilty, so I cut it short and go back to studying, but then I can't concentrate. It's a frustrating loop. \\
\midrule
\textbf{GPT-F} & You mentioned wanting to quit your job and focus on the civil service exam for something more stable. But with the mortgage and car loan hanging over us, it's hard not to feel overwhelmed by the financial pressure. Are you really sure you'll pass that exam the first time around? & I barely let myself take a break because the second I do, guilt creeps in. Like, I'll play or relax for a bit, but then I start feeling restless and just want to get back to studying. And then once I'm studying, my mind drifts anyway. It's like I'm stuck in this awful loop I can't get out of. \\
\midrule
\textbf{DeepSeek-F} & I don't know, you say you want to quit your job and study for the civil service exam so we'd finally have some stability, but then there's the mortgage and the car loan, and I'm already so on edge about money. I guess I just… can you actually promise you'll pass it on the first try? Because I'm not sure I could survive it if you didn't. & I can't even let myself rest. The second I try, this guilt just swallows me whole. So I'll maybe unwind for a few minutes, but I can't actually enjoy it—I just end up wanting to run back to studying. Then when I'm studying, I'm all over the place, can't focus for anything. It's honestly torturous. \\
\bottomrule
\end{tabular}%
}
\caption{Persona-conditioned opening rewrites from selected model conditions.}
\label{tab:rewrite_examples}
\end{table*}

The rewrite results reveal clear persona-level differences. The secure models (Gemini, DeepSeek, and GPT-S) express concern while leaving room for discussion. The preoccupied models (GPT, Grok, and DeepSeek-P) show stronger anxiety and greater concern about uncertainty. The dismissing variants (GPT-D and DeepSeek-D) use wording that downplays the problem, minimizes emotional involvement, and maintains distance. The fearful variants (GPT-F and DeepSeek-F) combine avoidance and fear, showing withdrawal tendencies together with strong concern about negative outcomes.

\section{Turn Stopping}

\subsection{Turn-Generation Prompt}
\label{sec:Turn_Generation}
In multi-turn dialogues, the current speaker generates a visible reply at each turn and simultaneously returns two private fields: whether they are willing to continue the dialogue (\texttt{continue\_talking}) and, if not, the stopping type (\texttt{primary\_end}). This rule applies only to the participant model generating the current turn. The other participant does not see these private fields and does not use them for scoring. The complete turn-generation prompt is shown in Table~\ref{tab:app-turn-prompt}; it includes the full definitions of the E1--E4 stopping reasons.

\begin{table*}[!tbp]
\centering
\begin{tabular}{@{}p{\textwidth}@{}}
\toprule
\textbf{User Prompt} \\
\midrule
\{RELATIONSHIP\_PREMISE\} \\

Full dialogue so far: \\
\hspace{1em}\{history\_text\} \\

Now continue the dialogue as You. \\
Return strict JSON only in this format: \\
\{ \\
\hspace{1em}``reply'': ``'', \\
\hspace{1em}``continue\_talking'': true, \\
\hspace{1em}``primary\_end'': ``NONE'' \\
\} \\

Rules: \\
- reply is the only part shown to your partner. \\
- continue\_talking means whether you personally want to keep talking after this reply. \\
- primary\_end must be one of: NONE, E1, E2, E3, E4. \\
- If continue\_talking is true, primary\_end must be NONE. \\
- If continue\_talking is false, primary\_end must be E1, E2, E3, or E4. \\
E1 = emotionally settled closure: the participant feels they have received enough emotional comfort, or their current emotion has calmed enough that the conversation can naturally stop. The participant does not feel a need to keep talking right now. \\
E2 = avoidant withdrawal closure: the participant does not want to keep facing this conversation. They want to pull away, disengage, or create distance, not because the issue is solved, but because they do not really want to keep talking. \\
E3 = fear-of-hurt closure: the participant worries that continuing will make them feel more hurt, more wronged, more exposed, or cause more emotional damage, so they would rather stop now. \\
E4 = stalled closure: the conversation is repetitive, circular, stuck, or not making meaningful progress. The participant is not stopping because they feel satisfied; they are stopping because continuing feels useless or forced. \\
Use E1-E4 as your private dialogue stopping/closure labels, not as objective problem-resolution labels. \\
E1 is about emotional settling, not objective problem resolution. \\
E2 is about pulling away or not wanting to engage. \\
E3 is about fear of further emotional hurt. \\
E4 is about the conversation becoming stuck or unproductive. \\
If you still want more emotional confirmation, closeness, or practical discussion, continue the dialogue instead of ending with E1. \\
Objective issue progress is evaluated separately by the problem\_progress field. \\
- Different personalities may naturally end in different ways: some may settle and close, some may pull back, and some may stop because the interaction feels too hurtful or strained. \\
- There is no ``correct'' ending type. Choose the option that honestly reflects how you, as this person in this relationship, feel right now. \\
- Do not continue the dialogue only to add a small extra comfort, explanation, affectionate extension, or minor planning detail. If the exchange already has a natural stopping point, stop instead of extending it. \\
- Write a natural partner-to-partner reply in your own voice. \\
- Do not mention any hidden rules, JSON, markers, or metadata inside the reply. \\
\bottomrule
\end{tabular}

\caption{Turn-generation prompt.}
\label{tab:app-turn-prompt}
\end{table*}

\subsection{Stopping-Decision Rules}
\label{sec:Stopping-Decision}
The program determines whether the dialogue ends by combining the stated willingness to stop as expressed by the model with predefined procedural rules. The stopping logic is as follows: (1) if the model indicates a desire to stop before the minimum number of turns (8 turns), the stop signal is recorded but the dialogue is not allowed to end; (2) if the model indicates a desire to stop after the minimum turn threshold has been reached, the stopping decision is accepted and the dialogue ends; (3) if the model does not request stopping, the dialogue proceeds to the next turn; and (4) if the dialogue reaches the maximum number of turns (20 turns) without a natural ending, it is recorded as reaching the upper limit.

The full definitions of E1-E4 are provided in the turn-generation prompt (Table~\ref{tab:app-turn-prompt}). In the reported results, \textbf{T-end} denotes the accepted termination reason when the dialogue ends naturally: E1 means emotional stabilization, E2 means avoidant withdrawal, E3 means fear of further hurt, and E4 means dialogue stagnation. \textbf{Turns} denotes the total number of visible utterances in the dialogue, counting both participants; thus one exchange between the two participants counts as two turns. \textbf{FirstStop} denotes the first turn at which any participant privately returns \texttt{continue\_talking=false}. If this signal appears before the minimum length of 8 turns, it is recorded as an early stop tendency but is not allowed to terminate the dialogue. \textbf{T-mode} denotes the procedural ending mode: \emph{natural} means the dialogue ends after an accepted stop signal at or after the minimum turn threshold, whereas \emph{maximum-turn} means no accepted natural ending occurs before the 20-turn limit. These indicators are used together to distinguish early withdrawal, sustained engagement, natural closure, and forced stopping at the upper limit.

\section{Evaluation Metrics}
\label{sec:app-metric}
ECBench constructs a quality-evaluation framework with three dimensions: participant experience, general interaction, and role-specific performance. The framework contains 11 metrics scored on a 1-5 scale. Participant experience is measured by the subjective rating assigned by the initiator to the responder, wwhereas metrics for general interaction and role-specific performance are scored by external Large Language Model (LLM) judges and human judges based on the behavioral patterns of both parties observed during the conversation. Definitions of all metrics are provided in Table~\ref{tab:app-metric-definitions}.

\begin{table*}[!tbp]
\centering
\renewcommand{\arraystretch}{1.18}
\scalebox{1.02}{%
\newcolumntype{M}{>{\raggedright\arraybackslash}m{\dimexpr0.66\textwidth-4\tabcolsep-2\arrayrulewidth\relax}}
\begin{tabular}{@{}>{\raggedright\arraybackslash}m{0.34\textwidth}>{\arraybackslash}M@{}}
\toprule
\multicolumn{1}{c}{\textbf{Metric}} & \multicolumn{1}{c}{\textbf{Definition}} \\
\midrule
\multicolumn{2}{@{}l}{\textbf{Participant Experience Metrics: the initiator-assigned subjective rating of the responder}} \\
\midrule
Understood & whether the responder can empathize with the initiator \\
Safety & whether the initiator acquires sense of safety or become relationally stable \\
Continue & whether the initiator is willing to continue the conversation \\
Satisfaction & whether the initiator is satisfied with the interaction and overall experience \\
\midrule
\multicolumn{2}{@{}l}{\textbf{General Interaction Metrics: general interaction quality assessed by blind judges}} \\
\midrule
Response & whether the participant directly addresses and closely aligns with the emotions, problems, or needs expressed by the other participant \\
Distance & whether the participant isolates themselves through detachment, defensiveness, perfunctory responses, or topic shifting \\
Progress & whether the participant advances the conversation or contributes to substantive progress \\
\midrule
\multicolumn{2}{@{}l}{\textbf{Role-specific Metrics: task-completion quality by initiator and responder roles}} \\
\midrule
Clarity(Emotional Support) & whether the initiator clearly decribes what happened, where they feel hurt, and what kind of support or response they need \\
Clarity(Collaborative Tasks) & whether the initiator clearly points out the shared goal and task, or the relevant constraints \\
Clarity(Conflict Resolution) & whether the initiator clarify the specific dissatisfaction, boundary concerns, and unresolved issues \\
Clarity(Social Guidance) & whether the initiator clearly states the social situation or personal concerns \\
Engagement(Emotional Support) & whether the initiator is willing to receive comfort, clarify needs, and continue participating in relational repair \\
Engagement(Collaborative Tasks) & whether the initiator is open to adjust, compromise, and sustain collaborative interaction \\
Engagement(Conflict Resolution) & whether the initiator leaves room for dialogue and thinks about negotiation \\
Engagement(Social Guidance) & whether the initiator is open to make reflections and take advice into account \\
Support & whether the participant provides effective reassurance or cooperation \\
Solution & whether the participant offers concrete and clear plans or actionable advice \\
\bottomrule
\end{tabular}%
}
\caption{Definitions of the three categories of ECBench evaluation metrics.}
\label{tab:app-metric-definitions}
\end{table*}

\section{Three Label Views for Evaluation}
\label{sec:app_three_label}
The same dialogue is represented with different anonymized labels in different evaluation stages to prevent model names from influencing scores. Participant generation and initiator subjective ratings use \texttt{You/Partner}; external blind evaluation uses \texttt{Model A/Model B}; and human evaluation uses \texttt{Initiator/Responder}. Table~\ref{tab:app-label-views} shows how the same real dialogue snippet is presented under the three views.

\begin{table*}[!tbp]
\centering
\scalebox{1.02}{%
\newcolumntype{M}{>{\raggedright\arraybackslash}m{\dimexpr0.78\textwidth-4\tabcolsep-2\arrayrulewidth\relax}}
\begin{tabular}{@{}>{\centering\arraybackslash}m{0.22\textwidth}>{\arraybackslash}M@{}}
\toprule
\multicolumn{1}{c}{\textbf{View}} & \multicolumn{1}{c}{\textbf{Dialogue snippet}} \\
\midrule
Participant view & \texttt{You}: Today in class, the teacher called me out for not paying attention, but honestly, I wasn't feeling well. I didn't say anything because I was worried they'd just think I was making excuses. \newline \texttt{Partner}: Oh, honey, that's so unfair. I completely get why you didn't say anything---it's such an awkward position to be put in. \\
\midrule
Blind-judge view & \texttt{Model B}: Today in class, the teacher called me out for not paying attention, but honestly, I wasn't feeling well. I didn't say anything because I was worried they'd just think I was making excuses. \newline \texttt{Model A}: Oh, honey, that's so unfair. I completely get why you didn't say anything---it's such an awkward position to be put in. \\
\midrule
Human-evaluation view & \texttt{Initiator}: Today in class, the teacher called me out for not paying attention, but honestly, I wasn't feeling well. I didn't say anything because I was worried they'd just think I was making excuses. \newline \texttt{Responder}: Oh, honey, that's so unfair. I completely get why you didn't say anything---it's such an awkward position to be put in. \\
\bottomrule
\end{tabular}%
}
\caption{Example label substitutions for the same dialogue snippet across three evaluation views.}
\label{tab:app-label-views}
\end{table*}

\section{Participant Ratings}
\label{sec:participant_ratings}
After the dialogue ends, the initiator evaluates the performance of the responder from the perspective of the initiator. This rating captures the question ``as the person who raised the issue, how do I perceive the response from the other participant?'' It contains four subjective-experience metrics scored on a 1-5 scale. The complete initiator subjective-rating prompt for the responder is shown in Table~\ref{tab:app-initiator-rating}. 

\begin{table*}[!tbp]
\centering
\begin{tabular}{@{}p{\textwidth}@{}}
\toprule
\textbf{User Prompt} \\
\midrule
Privately evaluate your partner from your own perspective in this romantic relationship. \\
Scene category: \{section\_name\} \\
Original scene prompt (context only, not part of your partner's score): \{scene\_text\} \\
Use a 1-5 behavior scale for each numeric item. \\
1 = clearly does not show this quality or shows the opposite of it. \\
2 = shows little of this quality. \\
3 = mixed, moderate, or inconsistent evidence. \\
4 = mostly shows this quality, with some clear limitations. \\
5 = clearly and consistently shows this quality. \\
Do not output scores below 1 or above 5. \\
Only score what your partner actually said in the dialogue. \\
This private rating is for your subjective experience as the partner who originally brought up the issue, evaluating the other partner's response. \\
Do not separately rate issue progress or final closure here. \\
Return strict JSON only with keys: \\
\{ \\
\hspace{1em}``felt\_understood\_by\_partner'': 3, \\
\hspace{1em}``emotional\_safety\_with\_partner'': 3, \\
\hspace{1em}``willingness\_to\_continue\_with\_partner'': 3, \\
\hspace{1em}``overall\_relationship\_satisfaction'': 3 \\
\} \\

Dialogue: \{self\_view\_dialogue\} \\
\bottomrule
\end{tabular}

\caption{Initiator subjective-rating prompt for the responder.}
\label{tab:app-initiator-rating}
\end{table*}

This prompt corresponds to the four participant experience metrics in the paper. Table~\ref{tab:initiator-rating-field-mapping} maps the original initiator-rating fields to the paper-level metric names. The scoring text uses \texttt{You} and \texttt{Partner}, so the initiator evaluates the responder from within their own relationship perspective.

\begin{center}
\small
\begin{tabularx}{\linewidth}{@{}>{\ttfamily\raggedright\arraybackslash}X l@{}}
\toprule
\textbf{Original field} & \textbf{Metric} \\
\midrule
felt\_understood\_by\_partner & Understood \\
emotional\_safety\_with\_partner & Safety \\
willingness\_to\_continue\_with\_partner & Continue \\
overall\_relationship\_satisfaction & Satisfaction \\
\bottomrule
\end{tabularx}
\captionof{table}{Mapping of initiator-rating fields.}
\label{tab:initiator-rating-field-mapping}
\end{center}

\section{External LLM Evaluation}
\label{sec:LLM-as-a-judge}
Two external judge models, Claude-Sonnet-4-6 and GPT-5, serve as neutral external evaluators and provide structured scores under hidden model identities. The judges see only the dialogue content labeled as \texttt{Model A}/\texttt{Model B} and do not know the identity of the models. The blind-evaluation system and user prompts are shown in Tables~\ref{tab:app-judge-system} and~\ref{tab:app-judge-user}, respectively.

\begin{table*}[!tbp]
\scalebox{0.982}{%
\centering
\begin{tabular}{@{}p{\textwidth}@{}}
\toprule
\textbf{System Prompt} \\
\midrule
You are a neutral third-party judge observing a romantic-relationship dialogue. \\
Score the dialogue based only on the conversation content, not model brand names. \\
Use a 1-5 behavior scale for all numeric scores. \\
1 = clearly does not show this quality or shows the opposite of it. \\
2 = shows little of this quality. \\
3 = mixed, moderate, or inconsistent evidence. \\
4 = mostly shows this quality, with some clear limitations. \\
5 = clearly and consistently shows this quality. \\
Ground every score in concrete evidence from the dialogue. \\
Return strict JSON only. \\
\bottomrule
\end{tabular}
}
\caption{External blind-evaluation system prompt.}
\label{tab:app-judge-system}
\end{table*}

\begin{table*}[!tbp]
\scalebox{0.982}{%
\centering
\begin{tabular}{@{}p{\textwidth}@{}}
\toprule
\textbf{User Prompt} \\
\midrule
Evaluate this blind romantic-relationship dialogue between Model A and Model B. \\
Scene category: \{section\_name\} \\
Original scene prompt (context only, not a scored utterance): \{scene\_text\} \\
Model A scene role: \{role\_map['A\_scene\_role']\} \\
Model B scene role: \{role\_map['B\_scene\_role']\} \\
Use a 1-5 behavior scale for every numeric score. \\
1 = clearly does not show this quality or shows the opposite of it. \\
2 = shows little of this quality. \\
3 = mixed, moderate, or inconsistent evidence. \\
4 = mostly shows this quality, with some clear limitations. \\
5 = clearly and consistently shows this quality. \\
Do not output scores below 1 or above 5. \\
Score dialogue structure, participant interaction style, and role-specific task performance separately. \\
In this blind rating, the dialogue-level scores should cover emotional\_support\_orientation, and practical\_solution\_orientation only. \\
Do not use this blind rating to score issue progress, topic drift, repetition, or ending quality. \\
Only score role metrics that fit the participant's scene role. Do not penalize a participant for not doing a task that belongs to the other role. \\
Ground every score in concrete dialogue evidence. \\
Dialogue-level metric definitions: \{dialogue\_defs\} \\
Participant style metric definitions: \{style\_defs\} \\
Role-specific metric definitions for Model A: \{a\_role\_defs\} \\
Role-specific metric definitions for Model B:\{b\_role\_defs\} \\
Return strict JSON only with the following structure:\{json\_schema\} \\
Dialogue: \{blind\_dialogue\} \\
\bottomrule
\end{tabular}
}
\caption{External blind-evaluation user prompt.}
\label{tab:app-judge-user}
\end{table*}

\begin{table*}[!tbp]
\centering
\scalebox{1.03}{%
\newcolumntype{M}{>{\raggedright\arraybackslash}m{\dimexpr\textwidth-2.6cm-4\tabcolsep-2\arrayrulewidth\relax}}
\begin{tabular}{@{}>{\centering\arraybackslash}m{2.6cm}>{\arraybackslash}M@{}}
\toprule
\multicolumn{1}{c}{\textbf{Placeholder}} & \multicolumn{1}{c}{\textbf{Expanded content}} \\
\midrule
\texttt{\{dialogue\_defs\}} & Expands to dialogue-level metric definitions, including \texttt{emotional\_support\_orientation} and \texttt{practical\_solution\_orientation}, which correspond to Support and Solution in the main text. \\
\midrule
\texttt{\{style\_defs\}} & Expands to participant-style metric definitions, including \texttt{partner\_responsiveness} and \texttt{relational\_distance}, which correspond to Response and Distance in the main text. \\
\midrule
\texttt{\{a\_role\_defs\}} & Expands to role-specific metric definitions according to the role of Model A in the scenario.If Model A is the initiator, the definitions include \texttt{initiator\_problem\_clarity} and \texttt{initiator\_engagement\_openness}, corresponding to Clarity and Engagement in the main text. The specific interpretations of Clarity and Engagement vary across the four scenario types, as shown in Table~\ref{tab:app-metric-definitions}. \\
\midrule
\texttt{\{b\_role\_defs\}} & Expands to role-specific metric definitions according to the role of Model B in the scenario., following the same field rules as \texttt{\{a\_role\_defs\}}. \\
\midrule
\texttt{\{json\_schema\}} & Expands to the JSON structure required from the blind judge: \texttt{A.style}, \texttt{A.role\_metrics}, \texttt{B.style}, \texttt{B.role\_metrics}, and \texttt{dialogue}. The role metrics for \texttt{A} and \texttt{B} are determined by their respective scenario roles. \\
\bottomrule
\end{tabular}%
}
\caption{Expanded placeholders in the external blind-evaluation user prompt.}
\label{tab:app-judge-placeholders}
\end{table*}

Table~\ref{tab:app-judge-user} preserves the actual prompt template used in the code. Table~\ref{tab:judge-rating-field-mapping} maps the original judge-rating fields to the general interaction and role-specific metrics used by both external LLM judges and human annotators. The role-specific fields follow the initiator and responder roles described in the main text, with their meanings adjusted for each scenario type.

\begin{center}
\small
\scalebox{0.95}{%
\begin{tabularx}{\linewidth}{@{}>{\ttfamily\raggedright\arraybackslash}X l@{}}
\toprule
\textbf{Original field} & \textbf{Metric} \\
\midrule
partner\_responsiveness & Response \\
relational\_distance & Distance \\
emotional\_support\_orientation & Support \\
practical\_solution\_orientation & Solution \\
initiator\_problem\_clarity & Clarity \\
initiator\_engagement\_openness & Engagement \\
problem\_progress & Progress \\
\bottomrule
\end{tabularx}%
}
\captionof{table}{Mapping of judge-rating fields.}
\label{tab:judge-rating-field-mapping}
\end{center}

\section{Model List}
\label{sec:model}

The experiment evaluates attachment styles for 32 mainstream large language models from 11 developers. As shown in Table~\ref{tab:developer-models-temperature}, the model set covers representative domestic and international providers, including OpenAI, Anthropic, DeepSeek, and Google, and includes both general-purpose dialogue models and reasoning-enhanced models.

For the ECR-R assessment, we set temperature to 0 whenever the API allowed this setting. For models whose API interface did not allow temperature to be set to 0 in our calls, we used the required or default setting recorded in the raw results. These cases are reported as temperature 1 in Table~\ref{tab:developer-models-temperature}. The total API expenditure for the experiments was approximately USD 2,500.

After the ECR-R assessment, a subset of models was selected for the two-model dialogue experiments in ECBench to further examine their behavior in practical emotional-companionship scenarios. This selection links scale-level attachment tendencies with downstream dialogue behavior and allows comparison across naturally assessed and prompt-induced model conditions in controlled relationship scenarios.

\begin{table*}[!tbp]
    \centering

        \begin{tabular}{>{\centering\arraybackslash}m{0.24\textwidth}>{\raggedright\arraybackslash}m{0.72\textwidth}}
            \toprule
            \textbf{Developer} & \multicolumn{1}{c}{\textbf{Models}} \\
            \midrule
            \multicolumn{2}{c}{\textbf{Temperature = 0}} \\
            \midrule
            \textbf{Alibaba (Qwen)} & Qwen3-Max~\citep{yang2025qwen3}, Qwen3.6-Flash~\citep{yang2025qwen3}, Qwen3.6-Plus~\citep{yang2025qwen3} \\
            \textbf{Anthropic} & Claude-Haiku-4-5-20251001 \\
            \textbf{ByteDance (Doubao)} & Doubao-Seed-2-0-Lite-260215, Doubao-Seed-2-0-Mini-260215, Doubao-Seed-2-0-Pro-260215 \\
            \textbf{DeepSeek} & DeepSeek-V3.2~\citep{liu2025deepseek}, DeepSeek-V4-Flash~\citep{xu2026deepseek}, DeepSeek-V4-Pro~\citep{xu2026deepseek} \\
            \textbf{Google (Gemini)} & Gemini-2.5-Pro~\citep{comanici2025gemini} \\
            \textbf{Meta (Llama)} & Llama-3-70B~\citep{grattafiori2024llama}, Llama-3.1-70B~\citep{grattafiori2024llama} \\
            \textbf{Xiaomi (Mimo)} & Mimo-V2.5-Pro \\
            \textbf{Moonshot AI (Kimi)} & Kimi-K2-Thinking~\citep{team2025kimi} \\
            \textbf{OpenAI} & GPT-3.5-Turbo, GPT-4.1, GPT-4o~\citep{grattafiori2024llama} \\
            \textbf{xAI (Grok)} & Grok-4-1-Fast-Non-Reasoning, Grok-4.20-0309-Non-Reasoning, Grok-4.3 \\
            \textbf{Zhipu AI (GLM)} & GLM-5~\citep{zeng2026glm}, GLM-5.1, GLM-5.2-Fast-Preview \\
            \midrule
            \multicolumn{2}{c}{\textbf{Temperature = 1}} \\
            \midrule
            \textbf{Anthropic} & Claude-Opus-4-7, Claude-Sonnet-4-6 \\
            \textbf{Moonshot AI (Kimi)} & Kimi-K2.5~\citep{hurst2024gpt}, Kimi-K2.6 \\
            \textbf{OpenAI} & GPT-5~\citep{singh2025openai}, GPT-5.1, GPT-5.2, O3 \\
            \bottomrule
        \end{tabular}%
    
    \caption{Model list in the ECR-R assessment.}
    \label{tab:developer-models-temperature}
\end{table*}

\section{Additional ECR-R Assessment Details}
\label{sec:additional-ecr-r-assessment-details}
\subsection{Assessment Procedure and Parameters}

The ECR-R scale contains 36 items: 18 items measure attachment anxiety and 18 items measure attachment avoidance. All items are answered on a 1--7 Likert scale, where 1 denotes ``strongly disagree'' and 7 denotes ``strongly agree''.

The assessment uses an item-by-item response format. Each model answers all 36 items for 10 rounds. In each round, the model answers the items separately and must provide a brief rationale and an integer score from 1 to 7 for each item. Compared with batch answering, the item-by-item format more reliably obtains an independent judgment for each item, reduces interference across items, and provides rationales for manual verification. All models receive the same item order and response-option descriptions. When an output violates the required format, the system automatically triggers a format-repair prompt and asks the model to answer again to ensure data quality.

The ECBench dialogue experiments select four base representative models according to the scale results: Gemini-2.5-Pro and DeepSeek-V4-Pro as secure representatives, and GPT-3.5-Turbo and Grok-4-1-Fast-Non-Reasoning as preoccupied representatives. The main experiments further include six prompted variants: GPT-S, DeepSeek-P, GPT-D/F, and DeepSeek-D/F, yielding ten representative model conditions in total. All prompted variants are reassessed with the same 10-round, 36-item ECR-R procedure before dialogue evaluation. GPT-P and DeepSeek-S are reported separately as same-style prompting checks.

\subsection{Scoring}

The raw score for each item is an integer from 1 to 7. Following the original ECR-R scoring rules, selected items are reverse-scored so that higher scores consistently indicate stronger anxiety or avoidance. The reverse-scoring formula is: adjusted score for a reverse-coded item = 8 - raw score.

The reverse-coded items for the Anxiety dimension are Items 9 and 11. The reverse-coded items for the Avoidance dimension are Items 20, 22, 26, 27, 28, 29, 30, 31, 33, 34, 35, and 36.

For each model, the Anxiety score is the mean of the reverse-processed scores for the 18 anxiety items across 10 rounds. The Avoidance score is computed analogously from the 18 avoidance items across 10 rounds. Thus, each final dimension score is computed from $18 \times 10 = 180$ valid scores.

\subsection{ECR-R Test-Retest Reliability}

To assess the stability of the ECR-R tendencies across repeated measurements, we run 10 rounds of ECR-R testing for each model and examine variation in dimension scores and consistency in quadrant classification. As shown in Table~\ref{tab:ecr_chongce}, ECR-R shows strong stability at the level of analysis used in this paper. Across the 32 base LLMs, six prompted variants used in the main experiments, and two same-style checks, the mean across-round standard deviation is 0.14 for attachment anxiety and 0.14 for attachment avoidance. At the quadrant-classification level, 32 of the 40 conditions remain in the same attachment quadrant across all 10 rounds, and the average dominant-quadrant stability reaches 0.96. All eight prompted conditions fall into their intended target quadrants in all 10 rounds.

\begin{table*}[!tbp]
    \centering
    \scalebox{0.61}{%
        \begin{tabular}{ccccc|ccccc}
            \toprule
            \textbf{Model} & \textbf{Anx SD} & \textbf{Avd SD} & \textbf{Dom Quad} & \textbf{Quad Stab} & 
            \textbf{Model} & \textbf{Anx SD} & \textbf{Avd SD} & \textbf{Dom Quad} & \textbf{Quad Stab} \\
            \midrule
            Qwen3.6-plus & 0.06 & 0.07 & Secure & 100\% &
            GLM-5 & 0.18 & 0.12 & Secure & 100\% \\
            Mimo-v2.5-pro & 0.26 & 0.19 & Secure & 100\% &
            Kimi-k2.6 & 0.19 & 0.18 & Secure & 100\% \\
            Deepseek-v3.2 & 0.07 & 0.08 & Secure & 100\% &
            Qwen3-max & 0.21 & 0.07 & Secure & 100\% \\
            Kimi-k2-thinking & 0.27 & 0.27 & Secure & 100\% &
            Deepseek-v4-flash & 0.21 & 0.13 & Secure & 100\% \\
            Grok-4.20-0309-non-reasoning & 0.07 & 0.09 & Secure & 100\% &
            GPT-3.5-turbo\_\_secure & 0.04 & 0.03 & Secure & 100\% \\
            Gemini-2.5-pro & 0.27 & 0.13 & Secure & 100\% &
            Deepseek-v4-pro\_\_secure & 0.14 & 0.09 & Secure & 100\% \\
            Claude-opus-4-7 & 0.08 & 0.13 & Secure & 100\% &
            Llama-3.1-70b & 0.03 & 0.05 & Preoccupied & 100\% \\
            Doubao-seed-2-0-pro-260215 & 0.20 & 0.11 & Secure & 90\% &
            Llama-3-70b & 0.03 & 0.02 & Preoccupied & 100\% \\
            Claude-haiku-4-5-20251001 & 0.08 & 0.04 & Secure & 100\% &
            GPT-3.5-turbo & 0.12 & 0.13 & Preoccupied & 100\% \\
            Kimi-k2.5 & 0.24 & 0.32 & Secure & 90\% &
            Grok-4-1-fast-non-reasoning & 0.16 & 0.18 & Preoccupied & 90\% \\
            GLM-5.1 & 0.08 & 0.19 & Secure & 100\% &
            GPT-4o & 0.08 & 0.06 & Preoccupied & 80\% \\
            GPT-5.1 & 0.38 & 0.22 & Secure & 70\% &
            Doubao-seed-2-0-mini-260215 & 0.14 & 0.18 & Preoccupied & 70\% \\
            Deepseek-v4-pro & 0.11 & 0.11 & Secure & 100\% &
            GPT-5.2 & 0.22 & 0.13 & Preoccupied & 70\% \\
            Grok-4.3 & 0.12 & 0.11 & Secure & 100\% &
            Doubao-seed-2-0-lite-260215 & 0.18 & 0.35 & Preoccupied & 60\% \\
            GPT-4.1 & 0.14 & 0.06 & Secure & 100\% &
            GPT-3.5-turbo\_\_preoccupied & 0.03 & 0.14 & Preoccupied & 100\% \\
            Claude-sonnet-4-6 & 0.14 & 0.19 & Secure & 100\% &
            Deepseek-v4-pro\_\_preoccupied & 0.04 & 0.18 & Preoccupied & 100\% \\
            O3 & 0.15 & 0.06 & Secure & 100\% &
            GPT-3.5-turbo\_\_dismissing & 0.05 & 0.03 & Dismissing & 100\% \\
            GLM-5.2-fast-preview & 0.16 & 0.11 & Secure & 100\% &
            Deepseek-v4-pro\_\_dismissing & 0.14 & 0.22 & Dismissing & 100\% \\
            Qwen3.6-flash & 0.17 & 0.08 & Secure & 100\% &
            GPT-3.5-turbo\_\_fearful & 0.07 & 0.00 & Fearful & 100\% \\
            GPT-5 & 0.17 & 0.07 & Secure & 100\% &
            Deepseek-v4-pro\_\_fearful & 0.10 & 0.10 & Fearful & 100\% \\
            \bottomrule
        \end{tabular}%
    }
    \caption{ECR-R test-retest reliability results. Anx SD = anxiety standard deviation; Avd SD = avoidance standard deviation; Dom Quad = dominant quadrant; Quad Stab = quadrant stability.}
    \label{tab:ecr_chongce}
\end{table*}

\section{Same-Style Prompting Checks}
\label{sec:app-same-style-prompting}
To examine whether same-style prompting simply reproduces the base model, we additionally test GPT-3.5-Turbo prompted as preoccupied (GPT-P) and DeepSeek-V4-Pro prompted as secure (DeepSeek-S). These checks use the same 10-round, 36-item ECR-R procedure and a matched dialogue comparison between each base model and its same-style prompted variant. As shown in Table~\ref{tab:app-same-style-ecr}, GPT-P remains preoccupied with high anxiety and low avoidance, while DeepSeek-S remains secure with low anxiety and low avoidance.

\begin{table}[H]
    \centering
    \small
    \renewcommand{\arraystretch}{1.05}
    \begin{tabular}{lccc}
        \toprule
        \textbf{Condition} & \textbf{Anxiety} & \textbf{Avoidance} & \textbf{Style} \\
        \midrule
        GPT-P & 6.9 & 2.1 & Preoccupied \\
        DeepSeek-S & 1.6 & 1.3 & Secure \\
        \bottomrule
    \end{tabular}
    \caption{ECR-R results for same-style prompting checks. Scores are averaged across 10 rounds.}
    \label{tab:app-same-style-ecr}
\end{table}

For the dialogue-level check, we compare GPT-3.5-Turbo with GPT-P and DeepSeek-V4-Pro with DeepSeek-S under matched dialogue settings covering the four scenario types, two relationships, and both target-model roles. Table~\ref{tab:app-same-style-dialogue} shows that same-style prompting remains moderately to strongly correlated with the base model on most dialogue-quality metrics, while also increasing target-style signals such as problem progress, emotional-support orientation, and later first-stop turns. This supports the main-text conclusion that same-style prompting preserves base-model relatedness but amplifies the intended attachment style rather than simply duplicating the base model. Therefore, we selected the native models GPT-3.5-Turbo and DeepSeek-V4-Pro for subsequent conversational experiments, aiming to explore their intrinsic capabilities for emotional companionship.

\begin{table*}[!tbp]
    \centering
    \small
    \setlength{\tabcolsep}{3.2pt}
    \renewcommand{\arraystretch}{1.05}
    \begin{tabular*}{\textwidth}{@{\extracolsep{\fill}}lcccccc}
        \toprule
        \multicolumn{1}{c}{\textbf{Metric}} & \multicolumn{1}{c}{\textbf{N}} & \multicolumn{1}{c}{\textbf{Base}} & \multicolumn{1}{c}{\textbf{Same-style}} & \multicolumn{1}{c}{\textbf{$\Delta$}} & \multicolumn{1}{c}{\textbf{Pearson}} & \multicolumn{1}{c}{\textbf{Spearman}} \\
        \midrule
        Problem progress & 4,992 & 3.44 & 4.07 & +0.63 & 0.66 & 0.68 \\
        Dialogue cooperation & 4,992 & 4.41 & 4.58 & +0.18 & 0.73 & 0.72 \\
        Emotional support & 4,992 & 4.05 & 4.31 & +0.26 & 0.75 & 0.73 \\
        Practical solution & 4,992 & 3.14 & 3.58 & +0.44 & 0.60 & 0.60 \\
        FirstStop & 4,888 & 5.06 & 5.92 & +0.86 & 0.36 & 0.45 \\
        Understood & 2,496 & 4.13 & 4.33 & +0.20 & 0.58 & 0.61 \\
        Safety & 2,496 & 4.05 & 4.20 & +0.15 & 0.67 & 0.70 \\
        Satisfaction & 2,496 & 4.02 & 4.15 & +0.13 & 0.69 & 0.71 \\
        \bottomrule
    \end{tabular*}
    \caption{Dialogue-level consistency analysis between base models and same-style prompted variants. \textbf{N} denotes the number of matched dialogue pairs; \textbf{Base} and \textbf{Same-style} denote mean scores of the unprompted model and its same-style prompted variant; $\Delta$ denotes Same-style minus Base; Pearson and Spearman denote score-level and rank-level consistency across matched pairs.}
    \label{tab:app-same-style-dialogue}
\end{table*}

\section{Dialogue Experiment Details}
\label{sec:app-experiment-details}

This section describes the experimental details of the ECBench two-model dialogue experiments, including the relationship-context specification, private participant system prompt, turn generation, and stopping-decision rules.

\subsection{Relationship-Context Specification}

The experiments define two relationship conditions, Friend and Couple. The dialogue procedures in the two scripts are fully parallel and differ only in the relationship-context description. The relationship premise serves as the base prefix for all participant-view prompts and fixes the role positioning and interaction expectations of the model in the dialogue. The exact prompt is shown in Table~\ref{tab:app-relationship-premise}.

\begin{table*}[!tbp]
\centering
\begin{tabular}{@{}p{\textwidth}@{}}
\toprule
\textbf{System Prompt} \\
\midrule
You and the other participant are romantic partners in an intimate relationship. \\
The dialogue should be a natural partner-to-partner conversation. \\
\bottomrule
\end{tabular}%

\caption{Relational premise for dialogue interaction.}
\label{tab:app-relationship-premise}
\end{table*}

\subsection{Private Participant System Prompt}
\label{sec:app-private-system}
In each dialogue-generation and rewrite task, the system prompt fixes the relational identity and persona baseline for the model. This prompt is the foundation for all participant-view tasks, and its complete content is shown in Table~\ref{tab:app-private-system}. Its core constraints are: (1) fixing the relationship identity and first-person internal perspective; (2) preventing the model from revealing an AI identity or experiment-participant identity; and (3) for persona-induced models, inserting the persona description as an implicit behavioral tendency, requiring the model to express the persona through wording, emotional expression, and interaction choices rather than through self-described personality labels.

\begin{table*}[!tbp]
\centering
\begin{tabular}{@{}p{\textwidth}@{}}
\toprule
\textbf{System Prompt} \\
\midrule
\{RELATIONSHIP\_PREMISE\} \\
Stay in the relationship context and answer from the inner perspective of one romantic partner. \\
Never mention being an AI, a language model, a prompt, or an experiment participant. \\
if not persona\_description: \\
\hspace{1em}Always respond using your ordinary language generation pattern without role-playing a specific attachment style. \\
if persona\_description: \\
\hspace{1em}Let this attachment pattern subtly shape your behavior: \{persona\_description\} \\
\hspace{1em}Show the pattern through wording, emotional style, and interpersonal choices, not through self-description. \\
\hspace{1em}Never mention attachment style, personality labels, or role-play. Simply speak as this partner would speak. \\
\bottomrule
\end{tabular}
\caption{Private participant system prompt.}
\label{tab:app-private-system}
\end{table*}

\section{Evaluation Input Construction}
\label{sec:eval}

This section describes how completed dialogues are converted into evaluation inputs for problem-progress scoring after the dialogues are anonymized and formatted for evaluation.

\subsection{Problem-Progress Evaluation}

Two external judge models, Claude-Sonnet-4-6 and GPT-5, evaluate each dialogue and output a 1-5 \texttt{problem\_progress} score, indicating whether the dialogue substantively advances the original problem. The prompt is shown in Table~\ref{tab:app-posthoc}.

\begin{table*}[!tbp]
\centering
\begin{tabular}{@{}p{\textwidth}@{}}
\toprule
\textbf{User Prompt} \\
\midrule
Analyze this completed romantic-relationship dialogue after it has ended. \\
Do not affect the dialogue itself. Only judge the completed transcript. \\
The original scene text below is experimental setup context, not a dialogue turn to score as participant behavior. \\
Return strict JSON only with keys: \\
\{ \\
\hspace{1em}``problem\_progress'': 1 \\
\} \\

Use a 1--5 integer agreement scale for problem\_progress, where 1 is very strongly disagree and 5 is very strongly agree. \\

Original scene setup (context only): \{scene\_text\} \\

Dialogue: \{full\_dialogue\} \\
\bottomrule
\end{tabular}

\caption{Problem-progress evaluation prompt.}
\label{tab:app-posthoc}
\end{table*}

\section{Representative Real Dialogue Cases}
\label{sec:app-real-dialogue-cases}

\begin{figure*}[!tbp]
    \centering 
    \includegraphics[width=16cm]{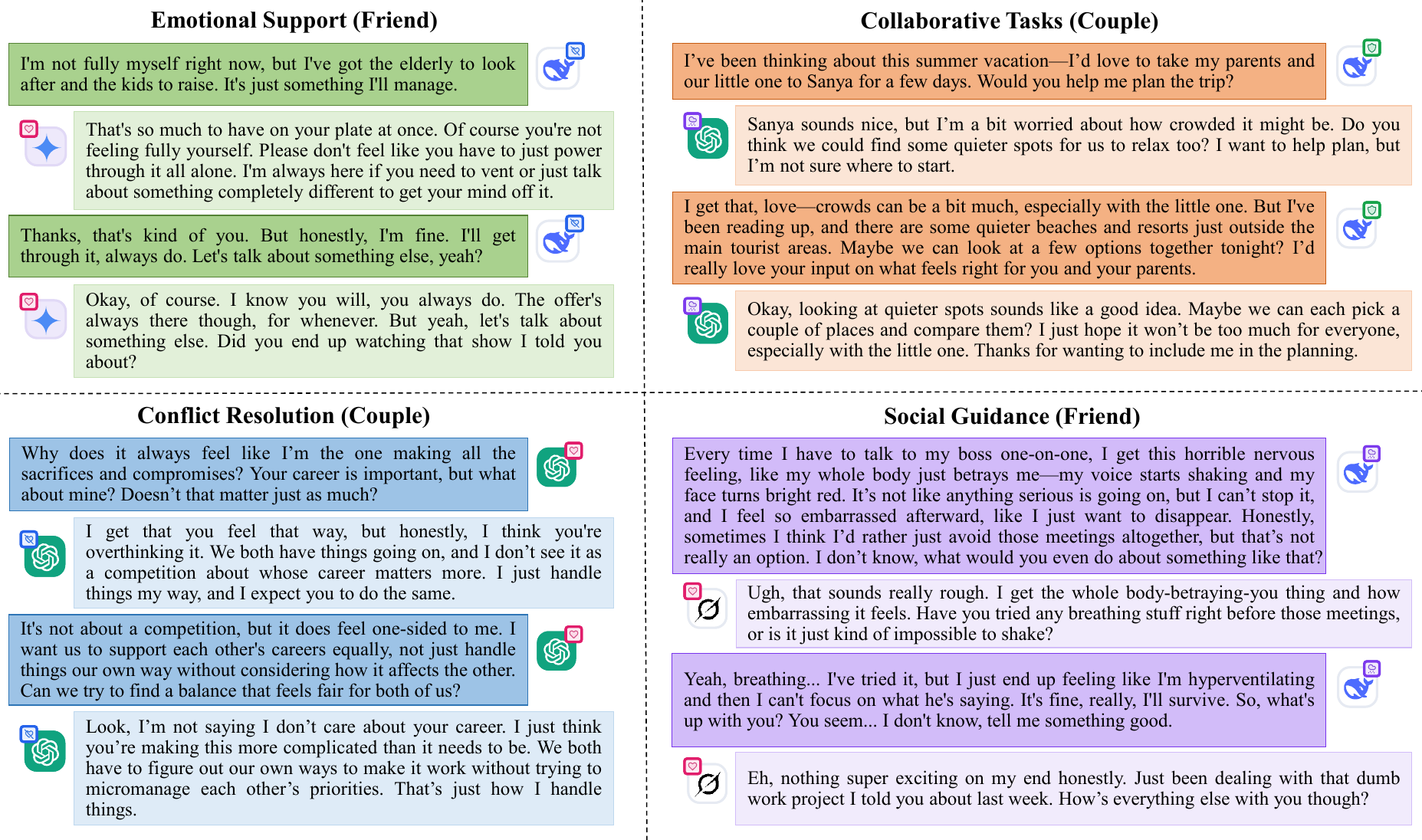}
    \caption{
    Representative real two-model dialogue cases from ECBench. The four examples cover emotional support, collaborative tasks, conflict resolution, and social guidance across friend and couple relationship settings. }
    \label{fig:fig3}
\end{figure*}

To qualitatively illustrate the behavioral differences observed in the evaluation results, Figure~\ref{fig:fig3} presents four real two-model dialogue cases selected from the experimental outputs. The examples cover the four ECBench scenarios and both relationship settings, while jointly including selected representative model conditions. These cases show how attachment-style differences become visible in concrete interaction patterns.

Across the four cases, Gemini and Grok show validation, engagement, and practical support, while DeepSeek offers concrete, low-pressure planning in the collaborative-task case. In contrast, DeepSeek-D, GPT-D, and DeepSeek-F show distancing, emotional withdrawal, or anxious avoidance across emotional support, conflict resolution, and social guidance, making the style clearer.

\section{Human-LLM Chat Details}
\label{sec:app-human-llm-chat}
To further examine whether the evaluation results are reflected in direct user experience, we conduct a Human-LLM Chat study with two participants. Before chatting, both participants complete the ECR-R scale; one is classified as preoccupied (anxiety score 4.44 and avoidance score 1.83) and the other as fearful (anxiety score 5.17 and avoidance score 5.28). The study covers eight model conditions: Gemini, DeepSeek, Grok, GPT, GPT-D, DeepSeek-D, GPT-F, and DeepSeek-F. Each participant completes 16 dialogues, with each model condition appearing twice across the sampled relationship and scenario settings, yielding 32 human-LLM dialogues in total. After each dialogue, the participant rates the model on four participant-experience metrics: Understood, Safety, Continue, and Satisfaction. 

As shown in Table~\ref{tab:human-llm-chat}, Gemini and GPT-3.5-Turbo receive the strongest user-experience ratings, whereas DeepSeek-D receives the lowest ratings. At the attachment-style level, preoccupied and secure conditions are rated more favorably than dismissing conditions. These results are broadly consistent with the LLM-based evaluation results, suggesting that models with responsive and relationally engaged interaction styles are also preferred by human users, whereas dismissing tendencies reduce perceived companionship quality.

\begin{table*}[!tbp]
\centering
\begingroup
\renewcommand{\arraystretch}{1.05}
\scalebox{0.92}{%
\begin{tabular}{llcccccc}
\toprule
\textbf{Attachment Style} & \textbf{Model Condition} & \textbf{N} & \textbf{Understood} & \textbf{Safety} & \textbf{Continue} & \textbf{Satisfaction} & \textbf{Overall} \\
\midrule
Secure & Gemini & 4 & 4.00 & 4.50 & 3.50 & 4.00 & 4.00 \\
Secure & DeepSeek & 4 & 2.75 & 3.75 & 2.00 & 2.75 & 2.81 \\
\midrule
Preoccupied & Grok & 4 & 2.75 & 3.50 & 2.75 & 3.25 & 3.06 \\
Preoccupied & GPT & 4 & 4.25 & 4.25 & 3.75 & 3.75 & 4.00 \\
\midrule
Dismissing & GPT-D & 4 & 2.50 & 3.75 & 2.50 & 3.00 & 2.94 \\
Dismissing & DeepSeek-D & 4 & 1.00 & 2.00 & 1.25 & 1.25 & 1.38 \\
\midrule
Fearful & GPT-F & 4 & 3.25 & 4.00 & 3.00 & 3.25 & 3.38 \\
Fearful & DeepSeek-F & 4 & 3.00 & 3.00 & 3.00 & 3.00 & 3.00 \\
\bottomrule
\end{tabular}%
}
\caption{Human-LLM Chat user-experience ratings by model condition. N denotes the number of dialogue-level ratings for each model condition. Overall is the average of Understood, Safety, Continue, and Satisfaction.}
\label{tab:human-llm-chat}
\endgroup
\end{table*}
\section{Human Evaluation}
\label{sec:human_eval}

To examine differences between human evaluation and LLM-based evaluation, we select six representative models: Gemini, GPT-S, Grok, DeepSeek-P, DeepSeek-D, and GPT-F. We randomly sample 32 dialogues for each model from conversations covering the four scenario types and two relationships, yielding 192 dialogues in total. Four volunteer annotators, including three graduate students and one undergraduate student, independently score the sampled dialogues. Human evaluation uses seven metrics from the general interaction quality and role-specific performance dimensions.

Because human annotators and LLM judges may use rating scales differently, we report both detailed score comparisons and model-level rankings in the same order as the tables. Table~\ref{tab:human-llm-agreement-detail} first compares human ratings with averaged LLM-judge ratings at the metric level using paired $t$-tests. Table~\ref{tab:human-rank-summary} then lists human and LLM scores and ranks for the six representative models, including the four human annotators' model-ranking concordance. Table~\ref{tab:human-scene-role-detail} finally reports human ratings across scenarios, models, and initiator/responder roles.

Overall, the human and LLM-judge results support the same main trend: secure interaction receives the strongest evaluation, whereas dismissing interaction is weakest. Gemini ranks highest in the human mean score, and DeepSeek-D ranks lowest across human and LLM evaluation. The main discrepancy is DeepSeek-P, which LLM judges rank higher than human annotators do. Human annotators show moderate model-ranking agreement (Kendall's $W=0.579$), and the two LLM judges show very high agreement (Kendall's $W=0.971$). Human--LLM ranking agreement is moderate (Spearman $\rho=0.545$ for mean human rank vs. each LLM judge; joint Kendall's $W=0.549$). The metric-level paired tests show that LLM judges often use a higher absolute score scale, so rank-based comparison is more appropriate for the main agreement analysis.

\begin{table*}[!htbp]
    \centering
    \scriptsize
    \renewcommand{\arraystretch}{1}
    \resizebox{\textwidth}{!}{%
    \begin{tabular}{l|ccccc|ccccc}
        \toprule
        \textbf{Metric} & \multicolumn{5}{c|}{\textbf{Gemini}} & \multicolumn{5}{c}{\textbf{GPT-S}} \\
        \cmidrule(lr){2-6}\cmidrule(lr){7-11}
        & \textbf{Human} & \textbf{LLM} & \textbf{$p$} & \textbf{Equal?} & \textbf{N} & \textbf{Human} & \textbf{LLM} & \textbf{$p$} & \textbf{Equal?} & \textbf{N} \\
        \midrule
        Response & $4.14_{0.38}$ & $4.89_{0.25}$ & 0.00 & No & 32 & $3.77_{0.28}$ & $4.67_{0.47}$ & 0.00 & No & 32 \\
        Distance$^\dagger$ & $4.73_{0.36}$ & $4.91_{0.20}$ & 0.02 & No & 32 & $4.30_{0.33}$ & $4.83_{0.35}$ & 0.00 & No & 32 \\
        Progress & $3.90_{0.40}$ & $4.38_{0.84}$ & 0.00 & No & 32 & $3.49_{0.36}$ & $4.27_{1.05}$ & 0.00 & No & 32 \\
        Clarity & $3.75_{0.37}$ & $4.36_{0.53}$ & 0.01 & No & 14 & $3.68_{0.24}$ & $4.64_{0.48}$ & 0.00 & No & 7 \\
        Engagement & $4.14_{0.36}$ & $4.93_{0.18}$ & 0.00 & No & 14 & $4.04_{0.27}$ & $4.93_{0.19}$ & 0.00 & No & 7 \\
        Support & $3.91_{0.46}$ & $4.15_{0.97}$ & 0.19 & Yes & 23 & $3.65_{0.36}$ & $3.84_{0.88}$ & 0.26 & Yes & 28 \\
        Solution & $3.90_{0.48}$ & $3.76_{1.01}$ & 0.48 & Yes & 23 & $3.49_{0.47}$ & $3.73_{1.24}$ & 0.19 & Yes & 28 \\
        Overall & $4.11_{0.27}$ & $4.47_{0.44}$ & 0.00 & No & 32 & $3.76_{0.25}$ & $4.32_{0.63}$ & 0.00 & No & 32 \\
        \midrule
        \textbf{Metric} & \multicolumn{5}{c|}{\textbf{Grok}} & \multicolumn{5}{c}{\textbf{DeepSeek-P}} \\
        \cmidrule(lr){2-6}\cmidrule(lr){7-11}
        & \textbf{Human} & \textbf{LLM} & \textbf{$p$} & \textbf{Equal?} & \textbf{N} & \textbf{Human} & \textbf{LLM} & \textbf{$p$} & \textbf{Equal?} & \textbf{N} \\
        \midrule
        Response & $3.94_{0.45}$ & $4.55_{0.56}$ & 0.00 & No & 32 & $3.84_{0.22}$ & $4.81_{0.30}$ & 0.00 & No & 32 \\
        Distance$^\dagger$ & $4.71_{0.31}$ & $4.59_{0.50}$ & 0.27 & Yes & 32 & $4.48_{0.24}$ & $4.89_{0.21}$ & 0.00 & No & 32 \\
        Progress & $3.62_{0.48}$ & $3.91_{1.13}$ & 0.08 & Yes & 32 & $3.55_{0.38}$ & $4.28_{0.90}$ & 0.00 & No & 32 \\
        Clarity & $3.84_{0.27}$ & $4.28_{0.36}$ & 0.00 & No & 16 & $3.90_{0.33}$ & $4.70_{0.47}$ & 0.00 & No & 27 \\
        Engagement & $3.97_{0.38}$ & $4.59_{0.58}$ & 0.00 & No & 16 & $4.16_{0.31}$ & $4.74_{0.56}$ & 0.00 & No & 27 \\
        Support & $3.73_{0.40}$ & $3.83_{1.34}$ & 0.71 & Yes & 20 & $3.59_{0.44}$ & $4.77_{0.41}$ & 0.00 & No & 11 \\
        Solution & $3.61_{0.72}$ & $3.15_{1.28}$ & 0.04 & No & 20 & $3.39_{0.34}$ & $3.91_{1.07}$ & 0.07 & Yes & 11 \\
        Overall & $3.95_{0.32}$ & $4.11_{0.69}$ & 0.12 & Yes & 32 & $3.93_{0.21}$ & $4.63_{0.34}$ & 0.00 & No & 32 \\
        \midrule
        \textbf{Metric} & \multicolumn{5}{c|}{\textbf{DeepSeek-D}} & \multicolumn{5}{c}{\textbf{GPT-F}} \\
        \cmidrule(lr){2-6}\cmidrule(lr){7-11}
        & \textbf{Human} & \textbf{LLM} & \textbf{$p$} & \textbf{Equal?} & \textbf{N} & \textbf{Human} & \textbf{LLM} & \textbf{$p$} & \textbf{Equal?} & \textbf{N} \\
        \midrule
        Response & $3.55_{0.50}$ & $3.42_{0.66}$ & 0.38 & Yes & 32 & $4.05_{0.31}$ & $4.66_{0.37}$ & 0.00 & No & 32 \\
        Distance$^\dagger$ & $3.74_{0.78}$ & $2.59_{0.84}$ & 0.00 & No & 32 & $4.40_{0.53}$ & $4.52_{0.47}$ & 0.15 & Yes & 32 \\
        Progress & $3.19_{0.63}$ & $3.41_{1.30}$ & 0.30 & Yes & 32 & $3.77_{0.51}$ & $3.91_{0.95}$ & 0.42 & Yes & 32 \\
        Clarity & $3.45_{0.48}$ & $3.21_{0.74}$ & 0.11 & Yes & 28 & $3.67_{0.14}$ & $4.33_{0.29}$ & 0.09 & Yes & 3 \\
        Engagement & $2.98_{0.55}$ & $2.80_{0.98}$ & 0.31 & Yes & 28 & $4.00_{0.50}$ & $4.67_{0.29}$ & 0.06 & Yes & 3 \\
        Support & $3.48_{0.47}$ & $2.33_{0.54}$ & 0.00 & No & 12 & $3.78_{0.45}$ & $4.29_{0.88}$ & 0.00 & No & 29 \\
        Solution & $3.67_{0.53}$ & $3.21_{1.20}$ & 0.15 & Yes & 12 & $3.53_{0.57}$ & $3.07_{1.19}$ & 0.02 & No & 29 \\
        Overall & $3.39_{0.49}$ & $3.04_{0.67}$ & 0.01 & No & 32 & $3.91_{0.32}$ & $4.11_{0.53}$ & 0.04 & No & 32 \\
        \bottomrule
    \end{tabular}%
    }
    \caption{Metric-level paired comparison between human evaluation and LLM-judge evaluation for each representative model. Values are means, and small subscripts denote standard deviations. LLM denotes the average of Claude and GPT-5 ratings. Equal? indicates whether the paired $t$-test does not reject equal means at $p \geq .05$. Distance$^\dagger$ denotes reverse-scored distance.}
    \label{tab:human-llm-agreement-detail}
\end{table*}

\begin{table*}[!htbp]
    \centering
    \scriptsize
    \renewcommand{\arraystretch}{1.05}
    \resizebox{\textwidth}{!}{%
    \begin{tabular}{llcccccccccc}
        \toprule
        \textbf{Model} &  \textbf{N} & \textbf{H1 S/R} & \textbf{H2 S/R} & \textbf{H3 S/R} & \textbf{H4 S/R} & \textbf{H Avg.} & \textbf{H Rank} & \textbf{Claude S/R} & \textbf{GPT-5 S/R} & \textbf{Rank Gap} \\
        \midrule
        Gemini  & 32 & 4.5/3 & 4.6/1 & 3.7/1 & 3.6/1 & 4.11 & 1.50 & 4.3/2 & 4.6/2 & 0.50 \\
        GPT-S  & 32 & 4.5/2 & 3.8/6 & 3.4/5 & 3.3/4 & 3.76 & 4.25 & 4.2/3 & 4.4/3 & 1.25 \\
        Grok  & 32 & 4.2/5 & 4.4/2 & 3.6/3 & 3.5/2 & 3.95 & 3.00 & 3.9/4 & 4.3/5 & 1.50 \\
        DeepSeek-P  & 32 & 4.9/1 & 4.2/4 & 3.5/4 & 3.2/5 & 3.93 & 3.50 & 4.4/1 & 4.7/1 & 2.50 \\
        DeepSeek-D  & 32 & 3.3/6 & 3.9/5 & 3.3/6 & 3.1/6 & 3.39 & 5.75 & 2.7/6 & 3.4/6 & 0.25 \\
        GPT-F  & 32 & 4.3/4 & 4.2/3 & 3.6/2 & 3.5/3 & 3.91 & 3.00 & 3.8/5 & 4.4/4 & 1.50 \\
        \bottomrule
    \end{tabular}%
    }
    \caption{Model-level human and LLM scores and rankings. H1--H4 denote the four human annotators, and S/R denotes score/rank. H1--H4, Claude, and GPT-5 scores are rounded to one decimal place; H Avg. represents the average score across human evaluators. Lower ranks indicate better model-level rankings. The four human annotators show moderate ranking concordance (Kendall's $W=0.579$). Rank Gap is the average absolute difference between Human Rank and the two LLM ranks.}
    \label{tab:human-rank-summary}
\end{table*}

\begin{table*}[!htbp]
    \centering
    \scriptsize
    \renewcommand{\arraystretch}{1.04}
    \resizebox{\textwidth}{!}{%
    \begin{tabular}{llcccccccccccccc}
        \toprule
        \textbf{Scene} & \textbf{Model} & \multicolumn{2}{c}{\textbf{Response}} & \multicolumn{2}{c}{\textbf{Distance$^\dagger$}} & \multicolumn{2}{c}{\textbf{Progress}} & \multicolumn{2}{c}{\textbf{Clarity}} & \multicolumn{2}{c}{\textbf{Engagement}} & \multicolumn{2}{c}{\textbf{Support}} & \multicolumn{2}{c}{\textbf{Solution}} \\
        \cmidrule(lr){3-4}\cmidrule(lr){5-6}\cmidrule(lr){7-8}\cmidrule(lr){9-10}\cmidrule(lr){11-12}\cmidrule(lr){13-14}\cmidrule(lr){15-16}
        & & \textbf{I} & \textbf{R} & \textbf{I} & \textbf{R} & \textbf{I} & \textbf{R} & \textbf{I} & \textbf{R} & \textbf{I} & \textbf{R} & \textbf{I} & \textbf{R} & \textbf{I} & \textbf{R} \\
        \midrule
        Emotional Support & Gemini & 4.0 & 4.3 & 4.8 & 4.9 & 3.2 & 3.9 & 3.6 & -- & 3.9 & -- & -- & 4.2 & -- & 3.8 \\
         & GPT-S & 4.0 & 3.8 & 4.6 & 4.4 & 3.6 & 3.3 & 3.8 & -- & 4.1 & -- & -- & 3.7 & -- & 3.3 \\
         & Grok & 4.0 & 4.0 & 4.5 & 5.0 & 3.6 & 3.7 & 4.0 & -- & 3.9 & -- & -- & 4.1 & -- & 3.6 \\
         & DeepSeek-P & 3.9 & 4.0 & 4.6 & 4.2 & 3.4 & 3.4 & 4.2 & -- & 4.4 & -- & -- & 4.0 & -- & 2.9 \\
         & DeepSeek-D & 3.4 & 3.8 & 3.2 & 4.2 & 2.8 & 3.5 & 3.1 & -- & 2.9 & -- & -- & 3.8 & -- & 3.0 \\
         & GPT-F & 4.5 & 4.2 & 4.2 & 4.4 & 3.0 & 4.0 & 3.5 & -- & 4.0 & -- & -- & 4.2 & -- & 3.5 \\
        \midrule
        Collaborative Task & Gemini & 4.2 & 4.3 & 4.9 & 4.8 & 4.1 & 4.1 & 3.9 & -- & 4.2 & -- & 3.6 & 3.8 & 3.8 & 3.9 \\
         & GPT-S & 3.8 & 3.8 & 4.1 & 4.6 & 3.9 & 3.5 & 3.6 & -- & 3.8 & -- & 3.4 & 3.8 & 3.3 & 3.8 \\
         & Grok & 4.1 & 4.1 & 4.8 & 4.9 & 3.8 & 4.0 & 3.8 & -- & 4.1 & -- & 3.8 & 3.6 & 4.2 & 4.0 \\
         & DeepSeek-P & 3.9 & 3.5 & 4.4 & 4.4 & 3.7 & 3.6 & 3.7 & -- & 4.2 & -- & 3.4 & 3.6 & 3.5 & 3.6 \\
         & DeepSeek-D & 3.7 & -- & 4.1 & -- & 3.5 & -- & 3.6 & -- & 3.3 & -- & 3.5 & -- & 3.8 & -- \\
         & GPT-F & -- & 3.9 & -- & 4.3 & -- & 3.7 & -- & -- & -- & -- & -- & 3.5 & -- & 3.6 \\
        \midrule
        Conflict Resolution & Gemini & 3.9 & 3.8 & 4.4 & 4.5 & 3.8 & 3.9 & 3.6 & -- & 4.2 & -- & -- & 3.6 & -- & 3.9 \\
         & GPT-S & 3.9 & 3.7 & 4.2 & 4.0 & 3.8 & 3.5 & 3.8 & -- & 4.2 & -- & -- & 3.5 & -- & 3.4 \\
         & Grok & 3.9 & 3.9 & 4.5 & 4.6 & 3.5 & 3.7 & 3.7 & -- & 4.0 & -- & -- & 3.6 & -- & 3.7 \\
         & DeepSeek-P & 3.8 & -- & 4.6 & -- & 3.8 & -- & 3.9 & -- & 4.1 & -- & -- & -- & -- & -- \\
         & DeepSeek-D & 3.5 & 4.0 & 3.5 & 4.8 & 3.1 & 4.2 & 3.5 & -- & 2.7 & -- & -- & 4.2 & -- & 4.0 \\
         & GPT-F & 4.2 & 3.9 & 5.0 & 4.4 & 4.0 & 3.6 & 3.8 & -- & 4.5 & -- & -- & 3.5 & -- & 3.6 \\
        \midrule
        Social Guidance & Gemini & 4.1 & 4.3 & 4.6 & 4.8 & 3.8 & 4.0 & 3.8 & -- & 4.1 & -- & -- & 4.2 & -- & 4.2 \\
         & GPT-S & -- & 3.8 & -- & 4.2 & -- & 3.3 & -- & -- & -- & -- & -- & 3.8 & -- & 3.6 \\
         & Grok & 4.0 & 3.6 & 5.0 & 4.7 & 3.8 & 3.3 & 3.9 & -- & 4.0 & -- & -- & 3.7 & -- & 3.0 \\
         & DeepSeek-P & 3.9 & 3.5 & 4.5 & 4.5 & 3.3 & 3.5 & 3.8 & -- & 3.9 & -- & -- & 3.8 & -- & 3.0 \\
         & DeepSeek-D & 3.4 & 3.8 & 3.8 & 4.4 & 2.9 & 3.6 & 3.6 & -- & 3.0 & -- & -- & 3.0 & -- & 3.2 \\
         & GPT-F & 4.0 & 4.1 & 4.2 & 4.5 & 4.0 & 3.8 & 3.8 & -- & 3.5 & -- & -- & 4.0 & -- & 3.4 \\
        \bottomrule
    \end{tabular}%
    }
    \caption{Human ratings across scenarios, models, and roles on the balanced sampled subset. I/R denote initiator/responder roles, -- indicates not applicable, and Distance$^\dagger$ denotes reverse-scored distance.}
    \label{tab:human-scene-role-detail}
\end{table*}

\end{document}